\documentclass[5p,authoryear]{elsarticle}

\newenvironment{widetable}[1][tbp]{\begin{table*}[#1]}{\end{table*}}
\newenvironment{widefigure}[1][tbp]{\begin{figure*}[#1]}{\end{figure*}}

\usepackage{amsmath,amssymb}
\usepackage{booktabs}
\usepackage{tabularx}
\usepackage{array}
\usepackage{xcolor}
\usepackage{url}
\usepackage[hidelinks]{hyperref}
\usepackage{graphicx}
\usepackage{listings}
\usepackage{tikz}
\usepackage[section]{placeins}
\usetikzlibrary{arrows.meta,positioning}
\lstdefinestyle{apiblock}{
  language=Python,
  basicstyle=\footnotesize\ttfamily,
  keywordstyle=\bfseries,
  commentstyle=\itshape\color{black!60},
  showstringspaces=false,
  columns=fullflexible,
  keepspaces=true,
  frame=single,
  framesep=4pt,
  xleftmargin=6pt,
  xrightmargin=6pt,
  breaklines=true,
}
\newcolumntype{Y}{>{\raggedright\arraybackslash}X}

\journal{International Journal of Forecasting}
\biboptions{authoryear,round}

\newcommand{\artifactdoi}{10.6084/m9.figshare.34874388}
\newcommand{\artifacturl}{\href{https://doi.org/\artifactdoi}{\nolinkurl{https://doi.org/\artifactdoi}}}

\begin{document}
\begin{frontmatter}

\title{SynForecast: Controlled Synthetic Time Series for Forecasting Research}

\author[aff1]{Olivier Sprangers\corref{cor1}}
\ead{olivier@nixtla.io}

\author[aff1]{Cristian Challu}
\ead{cristian@nixtla.io}

\author[aff1]{Jan Rathfelder}
\ead{jan@nixtla.io}

\author[aff1]{Han Wang}
\ead{han@nixtla.io}

\author[aff1]{Max Mergenthaler Canseco}
\ead{max@nixtla.io}

\cortext[cor1]{Corresponding author.}

\affiliation[aff1]{organization={Nixtla},
  city={San Francisco},
  state={CA},
  country={United States}}

\begin{abstract}
Synthetic time series constitute a substantial share of the training corpora of several recent forecasting foundation models. Existing studies report benefits at small data budgets and modest synthetic proportions, but each evaluates individual combinations of model and generator. Studying these benefits systematically requires generators that support controlled configuration, perturbation, reproducibility and mechanism-level ablation. We review 11 existing generators and find gaps in their combined support for these capabilities.

In this paper, we present SynForecast, an open-source Python library that generates synthetic time series from 32 generators, 19 of which implement named mechanisms with a published source in econometrics, quantitative finance, statistical physics and point-process theory. Its design supports controlled experiments: perturbations and families of mechanisms can be varied while everything else is held fixed, output comes in the long format that forecasting libraries accept, and output does not depend on the number of parallel workers.

We evaluate SynForecast on M4 at five frequencies, using a neural network and a gradient-boosted ensemble. We find that a classical baseline attains the lowest error in nine out of ten frequency-history combinations. Among the learned models, a synthetic prior outperforms the same network trained from scratch at a fixed budget by up to 34\% when only a few dozen observations are available. However, it loses to the network trained from scratch at full history at every frequency, and none of the three rescalings of history we pre-specified predicts where the sign changes. At the hourly frequency, where the prior fails at every history, we find that the synthetic corpus lacks daily cycles, and adding them reduces zero-shot error by 43 to 63\% for both models. Fine-tuning the pretrained model outperforms training from scratch by up to about 15\%, but at short histories a fixed 300-step fine-tune is worse than the pretrained model used as is. Finally, removing the classical family of mechanisms costs about 2\% for both models, and every other family has a smaller effect outside the hourly frequency.
\end{abstract}


\end{frontmatter}

\section{Introduction}
\label{sec:intro}
\label{sec:introduction}

Synthetic data play an increasingly important role in training time series foundation models. Chronos augments public datasets with series sampled from Gaussian-process priors \citep{ansari2024chronos}, while TimesFM assigns synthetic series a 20\% sampling weight during pretraining \citep{das2024timesfm}. Chronos-2 relies entirely on synthetic data for its multivariate and covariate-informed training tasks \citep{chronos2}. Other models use synthetic data as their sole source of pretraining examples: ForecastPFN and TempoPFN are trained exclusively on generated series \citep{dooley2023forecastpfn,moroshan2025tempopfn}, and \citet{oreshkin2026sarsim} train neural forecasting models on approximately one billion simulated series, reporting zero-shot accuracy on public benchmarks that exceeds AutoARIMA and several foundation models. These developments motivate studying how the composition and properties of synthetic training data affect forecasting performance.

Existing studies suggest that the benefits of synthetic training data depend on the forecasting setting. ForecastPFN achieves the most MSE wins among the evaluated methods when competing models receive training budgets of 50--250 observations \citep{dooley2023forecastpfn}. In the Chronos-T5 Small ablation, the most consistent improvements occur with approximately 10\% synthetic data \citep{ansari2024chronos}, while TimesFM benefits at underrepresented granularities but shows little change on the hourly ETT datasets \citep{das2024timesfm}. Other studies report less favorable results. \citet{cazaux2026synthetic} find that synthetic augmentation increases error in 67\% of their full-data augmentation trials across five architectures, with substantial differences between architectures. \citet{kuvshinova2024synthesize} show that supervised learning from limited real observations can outperform zero-shot forecasting with models trained on larger synthetic corpora. However, these findings concern different combinations of generating processes, forecasting models and training protocols. Hence, it remains unclear when synthetic pretraining improves performance on a new dataset, in particular as the available history changes.

Controlled simulation is a natural way to investigate these dependencies, and has an established role in forecasting research \citep{kang2017instance,hewamalage2022global}. Answering when synthetic pretraining helps requires experiments that vary one factor at a time, such as the generating mechanisms, a perturbation or the forecasting model, while holding the others fixed. However, existing generators make such experiments difficult in four ways. First, several return only bare arrays without frequencies, time origins or series identifiers, such as \texttt{TimeSynth}\footnote{Throughout the paper, software packages are set in typewriter type (e.g., \texttt{SynForecast}, \texttt{pandas}), whereas models and algorithms are set in roman type (e.g., NHITS, KernelSynth).}, so researchers must reconstruct forecasting panels by hand, where alignment errors can go unnoticed. Second, where perturbations are available, their isolation from the base process is not guaranteed by design: whether the underlying series stays the same when an anomaly or changepoint is enabled depends on draw order or a shared random stream, so a stress test can confound the perturbation with a different base trajectory. Third, determinism under parallel generation is not guaranteed: most generators either generate in a single process, which limits scale, or give no guarantee that output is unchanged across worker counts. Fourth, most pretraining generators were built for a single study and implement a single generating family, so the contribution of a mechanism cannot be measured by removing it from a corpus.

We therefore identify four software requirements, one for each of these gaps: (i) long-format panel output with explicit series identifiers, time origins and frequencies; (ii) perturbations that can be varied while the underlying generating process remains fixed; (iii) identical output for a fixed seed and configuration regardless of the number of parallel workers; and (iv) diverse generating mechanisms organized into named families that can be selectively excluded. Of the 11 existing generators we examined, none fully satisfies more than two of these requirements (\ref{app:generators}). 

In this paper, we present \texttt{SynForecast}, an open-source Python library that implements these four requirements within a common interface for synthetic time series generation. We use it to examine how the benefits of synthetic pretraining vary with history length, forecasting model, and the composition of the synthetic corpus.

\emph{Contributions.} We make two main contributions:
\begin{enumerate}
    \item We introduce \texttt{SynForecast}, a software framework for controlled synthetic time series generation (Section~\ref{sec:design}). It combines long-format panel output, perturbations applied separately from the generating processes, and randomness sampled before parallel dispatch, so that output is reproducible across worker counts. Its 32 generators include 19 mechanisms drawn from published work, organized into named families that support systematic ablation.
    \item We study synthetic pretraining with two small forecasting models on five M4 frequencies (Section~\ref{sec:results}). We show that a synthetic prior outperforms training from scratch by up to 34\% at short histories but not at full history, and we test whether the number of observations, the number of seasonal cycles or the ratio of history length to forecast horizon explains where this transition occurs. We also measure the contribution of individual generator families across models, trace the failure of the prior at the hourly frequency to missing daily cycles, and use controlled perturbations to assess the effect of structural breaks on three forecasting methods.
\end{enumerate}

\emph{Scope.} We use small, open models so that the experiments can be reproduced on modest hardware. Hence, our findings characterize the tested models and configurations, and their applicability to large forecasting foundation models remains to be evaluated. \texttt{SynForecast} also supports data augmentation (Section~\ref{sec:augmentation}), but we leave its evaluation, including a matched comparison of bootstrap and parametric augmentation, to future work.

\texttt{SynForecast} is released under the Apache 2.0 license at \url{https://github.com/Nixtla/synforecast}, with documentation at \url{https://nixtlaverse.nixtla.io/synforecast/}.

\section{Background and related work}
\label{sec:related}

We review three lines of work that generate synthetic time series: for pretraining foundation models, for benchmarking forecasting methods, and for augmenting observed series. For each line of work, we position \texttt{SynForecast} relative to existing approaches. We refer the interested reader to the survey of \citet{liu2025synthetic} for a broader overview of synthetic data for time series foundation models, and to \citet{petropoulos2022forecasting} for the wider forecasting literature. Our focus is on generation from explicitly specified processes. Learned generative models, including diffusion models \citep{yuan2024diffusionts} and the methods evaluated in TSGBench \citep{ang2023tsgbench}, instead estimate generative distributions from observed data. These methods can support several downstream applications, but a detailed review is outside the scope of this paper.

\subsection{Synthetic data for pretraining}

Synthetic generation provides a scalable way to expand the volume and diversity of forecasting training corpora. Existing approaches differ in their generating assumptions and how synthetic series are incorporated into training. Chronos samples from Gaussian-process priors over composed kernels \citep{ansari2024chronos}; TimesFM mixes about three million series built from ARMA processes, seasonal patterns, trends and step functions into a real corpus that includes M4 \citep{das2024timesfm}; and ForecastPFN \citep{dooley2023forecastpfn}, Mamba4Cast \citep{bhethanabhotla2024mamba4cast} and TempoPFN train on synthetic data alone, the last from a released pipeline that includes stochastic differential equations, Gaussian processes and audio synthesis \citep{moroshan2025tempopfn}. SarSim0 \citep{oreshkin2026sarsim} superposes stable SARIMA trajectories into multi-seasonal traces, adds rate-based heavy-tailed noise and supports training on about one billion series generated on the fly. On the public GiftEval benchmark, models pretrained on SarSim0 data improve on AutoARIMA and recent foundation models \citep{oreshkin2026sarsim}, and TempoPFN surpasses most models trained on real data \citep{moroshan2025tempopfn}.

Three features of this literature matter for what follows. First, generators differ in breadth: KernelSynth and SarSim0 build primarily on Gaussian processes and SARIMA models, respectively, whereas TimesFM combines four types of process and the released TempoPFN code includes 14 named generators, which its paper groups into ten types. Second, several studies examine the contributions of individual generators or components. TempoPFN reports leave-one-generator-out ablations and finds that excluding its stochastic-differential-equation generator raises normalized CRPS on GIFT-Eval by about 26\% \citep{moroshan2025tempopfn}. The SarSim0 study compares training on SarSim0, KernelSynth and ForecastPFN data across the same forecasting models, and separately ablates components of the SarSim0 pipeline \citep{oreshkin2026sarsim}. Third, these pipelines were developed for particular pretraining studies, whereas \texttt{SynForecast} packages generation in a standalone library for reuse across forecasting experiments.

\subsection{Generation for benchmarking and evaluation}
\label{sec:gratis}

Synthetic generation also supports forecasting evaluation by providing data with controlled characteristics. Forecasting libraries such as \texttt{GluonTS} \citep{alexandrov2020gluonts} and \texttt{Darts} \citep{herzen2022darts} include utilities for generating artificial series, in their \texttt{dataset.artificial} and \texttt{utils.timeseries\_generation} modules, respectively. More extensive simulation studies investigate how forecasting performance depends on the generating process. \citet{hewamalage2022global}, for example, simulate autoregressive and seasonal autoregressive processes, SETAR models, logistic maps, and Mackey--Glass systems while varying series length, the number of series, heterogeneity, and model complexity to compare global and local forecasting methods. This empirical work complements theoretical results on the expressiveness and complexity of global forecasting models \citep{montero2021principles}.

GRATIS \citep{kang2020gratis} is a particularly relevant approach to reusable synthetic generation. It uses mixture autoregressive (MAR) models \citep{wongli2000mar}, which can capture heavy tails and heteroskedasticity, to generate series with diverse properties, including non-stationarity and multiple seasonalities. Its evaluation demonstrates broad coverage of benchmark datasets, including M4, in a projected feature space. GRATIS also supports generation with target characteristics by using a genetic algorithm to optimize MAR parameters, extending earlier work that directly evolved time series observations \citep{kang2017instance}.

\texttt{SynForecast} includes a GRATIS-inspired MAR generator, adapted rather than reproduced: it samples stationary autoregressive components, standardizes series by default, and targets four summary features. The two approaches differ in how a corpus is specified, not in how much of the feature space it can reach. GRATIS targets diversity and features within the MAR family, whereas \texttt{SynForecast} exposes named mechanisms whose contributions can be measured by changing the composition of the corpus (Section~\ref{sec:ablation}). Mechanism-based specification and feature-based evaluation are complementary, and we use both in Section~\ref{sec:hourly}.

Separating a base process from perturbations also has precedents, for example in \texttt{GutenTAG}, which configures anomalies separately from base oscillations \citep{gutentag}. \texttt{SynForecast} adopts this separation so that perturbations can vary while the base series stays fixed (Section~\ref{sec:overlays}).

\subsection{Augmentation of observed series}

Augmentation supplements observed datasets with additional series, often obtained through resampling, transformations, or simulation from fitted models. An established application is bagging exponential smoothing models fitted to series reconstructed from a moving block bootstrap of the STL remainder of Box--Cox transformed series \citep{bergmeir2016bagging}. \citet{petropoulos2018bagging} attribute much of the improvement to accounting for uncertainty in model selection. For global forecasting models, \citet{bandara2021improving} study augmentation through bootstrapping and dynamic time warping barycentric averaging \citep{petitjean2011global}, alongside transfer learning from GRATIS-generated series, and report accuracy gains in settings with limited training data.

Synthetic series also support model selection and pretraining. FFORMS augments the reference set used to train its model-selection classifier with series simulated from fitted models \citep{talagala2023metalearning}. Chronos uses TSMixup to construct convex combinations of real series; an ablation with Chronos-T5 Small finds similar in-domain accuracy and improved zero-shot accuracy \citep{ansari2024chronos}. However, these benefits vary across settings: \citet{cazaux2026synthetic} report that augmentation degrades accuracy in 67\% of their trials using the full real training set, with substantial variation across architectures.

\texttt{SynForecast} brings existing generating mechanisms and augmentation methods into a common interface for forecasting research. Its contribution is the software design that allows researchers to vary corpus composition and perturbations, reproduce the resulting data, and compare their effects across forecasting models.

\section{Design}
\label{sec:design}

In this section, we first state the four requirements that guide the design of \texttt{SynForecast}. Next, we describe the generators, the output format, the perturbation overlays and the seeding scheme that makes generation deterministic under parallelism. We end the section with augmentation and throughput.

\subsection{Four requirements for controlled forecasting experiments}

Four requirements guide the design of \texttt{SynForecast}, one for each gap identified in Section~\ref{sec:intro}: (i) forecasting-ready output, in which observations carry a frequency, a starting timestamp and a series identity \citep{wang2020tsibble}; (ii) separable perturbations, so that a changepoint, an anomaly or a burst of missing values can vary while the base trajectory stays fixed; (iii) determinism under parallelism, so that a fixed configuration and seed produce the same corpus regardless of worker count or scheduling; and (iv) removable families of mechanisms. Comparing corpora with and without a family requires a defined protocol for corpus size, sampling weights and training budget, and measures the contribution of that family to a particular mixture.

Figure~\ref{fig:schematic} illustrates how \texttt{SynForecast} addresses the first three requirements, which Sections~\ref{sec:interface} to~\ref{sec:determinism} examine in turn. Section~\ref{sec:surface} describes the generator families that support the fourth. Augmentation introduces an additional concern when generators are fitted to observed data: preventing leakage requires fitting only on the training portion of each evaluation split. Section~\ref{sec:augmentation} discusses this protocol.

\begin{widefigure}[tbp]
\centering
\resizebox{\linewidth}{!}{\definecolor{nxink}{HTML}{222121}\definecolor{nxpaper}{HTML}{F6F4F0}
\begin{tikzpicture}[
  font=\footnotesize, node distance=3mm and 4.2mm,
  box/.style={draw=nxink, line width=0.5pt, fill=nxpaper, sharp corners, align=center,
              inner sep=3pt, minimum height=11mm},
  arr/.style={-{Stealth[length=1.5mm]}, draw=nxink, line width=0.5pt},
  req/.style={font=\scriptsize\itshape, text=nxink!70, align=center,
               execute at begin node={\hyphenpenalty=10000}},
]
\node[box] (cfg) {\textbf{Configuration}\\ $\theta$, overlays, $s$, $n$};
\node[box, right=of cfg] (rng) {\textbf{Master stream $R(s)$}\\
  $n_i,\ g_i,\ (c_i, a_i, m_i)$};
\node[box, right=of rng] (gen) {\textbf{Workers}\\ $x_i = G_\theta(g_i, n_i)$};
\node[box, right=of gen] (ov) {\textbf{Overlays}\\
  $y_i = M_{m_i}\circ A_{a_i}\circ C_{c_i}(x_i)$};
\node[box, right=of ov] (out) {\textbf{Long format}\\
  \texttt{unique\_id}, \texttt{ds}, \texttt{y}};
\draw[arr] (cfg) -- (rng);
\draw[arr] (rng) -- (gen);
\draw[arr] (gen) -- (ov);
\draw[arr] (ov) -- (out);
\node[req, below=of rng, text width=3.6cm] {(iii) every draw made before
  dispatch: output independent of the worker count};
\node[req, below=of ov, text width=3.4cm] {(ii) $x_i$ independent of the overlays;
  a disabled overlay is the identity};
\node[req, below=of out, text width=2.6cm] {(i) read directly by forecasting
  libraries};
\end{tikzpicture}
}
\caption{From configuration to panel. Before dispatch, a master random stream initialized with seed $s$ samples series lengths $n_i$, generation seeds $g_i$, and one seed per overlay and series. Workers generate base trajectories $x_i$ under mechanism configuration $\theta$ using their assigned generation seeds. Changepoints $C$, anomalies $A$ and missingness $M$ are then applied in sequence, each using its own seeded random stream. The resulting series are assembled into a long-format dataframe. Each stage is annotated with the design requirement it addresses.}
\label{fig:schematic}
\end{widefigure}

\subsection{The generator surface}
\label{sec:surface}

\texttt{SynForecast} exposes 32 generators organized into five groups:
\begin{itemize}
  \item \emph{Statistical (5):} random walk, seasonal, SARIMA, ETS and integer-valued autoregression.
  \item \emph{Stochastic (13):} GARCH, Ornstein--Uhlenbeck, geometric Brownian motion, jump diffusion, Poisson, cyclic, fractional Brownian motion, Hawkes, stochastic volatility, regime switching, chaotic systems, bounded processes and L\'evy processes. These draw mainly on quantitative finance, statistical physics and point-process theory.
  \item \emph{Multivariate (3):} copula, VAR and Gaussian process, with copula and VAR generators providing explicit dependence across series.
  \item \emph{Domain (7):} intermittent demand, energy load, IoT sensors, vital signs, clickstream, daily active users and state-space models.
  \item \emph{Pretraining (4):} TSI and TCM, used as univariate generators by \citet{chronos2}; KernelSynth, following \citet{ansari2024chronos}; and a GRATIS-inspired mixture-autoregressive generator (Section~\ref{sec:gratis}). These generators randomize process structure or component composition across series to produce diverse training corpora.
\end{itemize}

The first four groups contain 28 generators: 19 implement named mechanisms with published sources, four are textbook constructions and five are stylized domain simulators. The supplementary material describes the model or recipe implemented by each generator and documents its provenance.

All 32 generators share a configuration and generation interface through a common base class. The following example uses the default SARIMA parameters and a fixed random seed:

\begin{lstlisting}[style=apiblock]
from synforecast.generators import SARIMAGenerator

gen = SARIMAGenerator(
    min_length=365, max_length=365, freq="D", seed=42
)
df = gen.generate(n_series=1_000)
\end{lstlisting}

The call generates 1,000 daily SARIMA series, each containing 365 observations. The parameters \texttt{min\_length} and \texttt{max\_length} specify inclusive length bounds; when they differ, lengths are sampled uniformly, allowing panels with unequal series lengths. The \texttt{freq} parameter accepts \texttt{pandas} offset aliases, such as \texttt{"D"} for daily observations, or a positive integer specifying the spacing of an integer time index. Figure~\ref{fig:gallery} illustrates three generators from each group and the overlays described in Section~\ref{sec:overlays}.

\begin{widefigure}[tbp]
\centering
\includegraphics[width=\linewidth]{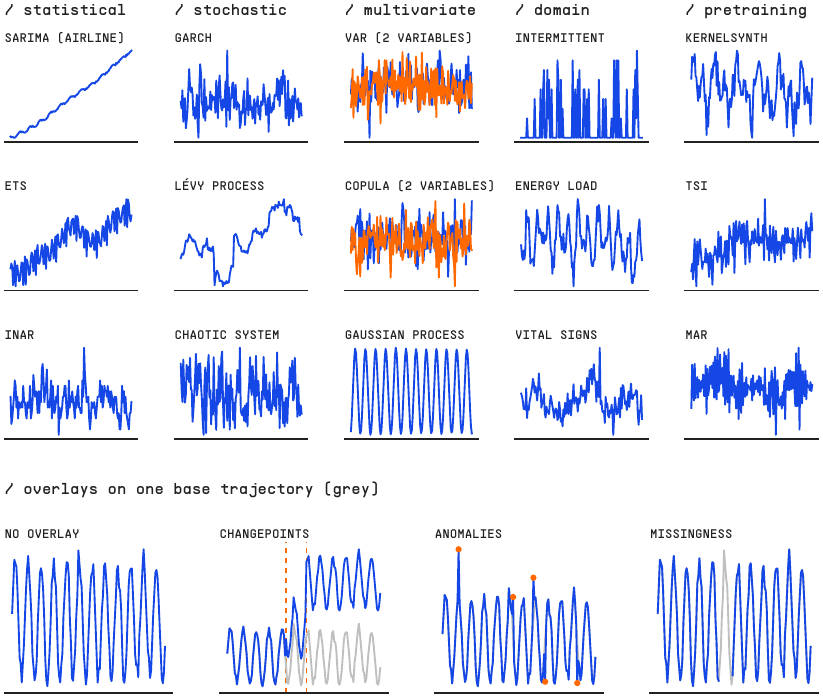}
\caption{Example output for monthly series of length 144. Top: three generators from each of the five groups, using default mechanism parameters except for the seasonal settings of SARIMA, ETS and the Gaussian process. Jointly generated panels are illustrated with two series each. Bottom: a seasonal series with each overlay enabled separately, superimposed on the same base trajectory without overlays (gray). Dashed lines indicate changepoint positions, markers identify anomalies, and the gray base trajectory remains visible through the missing block.}
\label{fig:gallery}
\end{widefigure}

\subsection{Output format}
\label{sec:interface}

\texttt{generate} returns a long-format dataframe with columns \texttt{unique\_id}, \texttt{ds} and \texttt{y}, following the schema used across the Nixtlaverse forecasting libraries. The default \texttt{pandas} output can therefore be passed to \texttt{StatsForecast}, \texttt{MLForecast} and \texttt{NeuralForecast} without restructuring the data. Column names are configurable at construction to accommodate other project conventions. Outputs that are only bare arrays, such as those of \texttt{TimeSynth} and the \texttt{TSGM} simulators, and the KernelSynth script, which attaches only a fixed dummy start date (\ref{app:generators}), require additional code to construct a forecasting panel with timestamps, frequencies, series identifiers and consistent ordering. \texttt{SynForecast} centralizes this step in a shared implementation. \texttt{gratis} already supports keyed temporal output in R; \texttt{SynForecast} adopts this approach for integration with Python forecasting libraries.

Schema compatibility does not replace data validation. Missing values may be intentional, as in sensor-failure simulations or missingness overlays, and their treatment depends on the downstream model and experimental protocol. Numerical divergence can also lead to non-finite values that require detection and handling. In our experiments, we remove series containing non-finite values before training. The common schema reduces the need for custom data preparation while leaving these validity checks explicit.

A shared output-construction path supports \texttt{pandas}, \texttt{Polars}, \texttt{cuDF}, \texttt{Modin} and \texttt{PyArrow}; the full test suite covers \texttt{pandas} and \texttt{Polars}, and the other three are smoke-tested. Users can select the representation that matches their downstream pipeline, subject to the formats accepted by the consuming library. \texttt{Polars} output is constructed without an intermediate \texttt{pandas} dataframe; selecting \texttt{cuDF} materializes the generated arrays on the GPU.

\subsection{Perturbations as overlays}
\label{sec:overlays}

Changepoints, anomalies and missingness are configured through the base class and applied after generation. All 32 generators therefore expose the same perturbation interface. Exogenous-variable options are configured through the same base class and add covariates or metadata to the output. The following example combines the three perturbation types:

\begin{lstlisting}[style=apiblock]
gen = SARIMAGenerator(
    min_length=365, max_length=365, freq="D", seed=42,
    changepoints=True, num_changepoints=3,
    changepoint_type="mixed",
    anomalies=True, anomaly_fraction=0.02,
    anomaly_types=["spike", "level_shift"],
    missing_data=True, missing_pattern="block",
    missing_block_size=7,
)
\end{lstlisting}

\noindent Each perturbation has explicit parameters. Changepoints support level, trend, variance and mixed changes, with a configurable count and optional relative locations and magnitudes. Missingness supports random, block and seasonal patterns, controlled by a rate and pattern-specific parameters such as block size or seasonal period. Anomalies have a configurable fraction, a selection of types, per-type magnitudes and a duration for temporary level shifts.

A separate option controls the innovation distribution in generators that use the shared noise sampler. The available distributions are normal, Student's $t$, Laplace, uniform and skew-normal, allowing noise tails and asymmetry to vary without introducing a new generator class. Additive anomaly magnitudes are expressed in the units of the series; the default spike-magnitude parameter is ten units. Consequently, a fixed setting has different relative effects across series scales. Comparisons intended to hold relative perturbation strength fixed must adjust these magnitudes accordingly, as in Section~\ref{sec:overlaydemo}.

The overlay implementation is shared across generator classes, avoiding separate implementations for each combination of mechanism and perturbation. For controlled comparisons, the essential property is that overlay settings do not affect the random stream used to generate the base trajectory. Initializing generators with the same base configuration and seed therefore preserves the underlying trajectory when perturbations are enabled or modified. This supports paired stress tests in which forecasting accuracy is compared on matched base data. Section~\ref{sec:determinism} describes the seed allocation that ensures this separation.

\subsection{Determinism under parallelism}
\label{sec:determinism}

Generation is parallel by default. The \texttt{n\_jobs} argument controls the worker count, with all available logical cores used unless the runtime configuration specifies otherwise. Work is dispatched to a Rust batch implementation built on \texttt{Rayon} or to a threaded implementation for mechanisms without a batch path. Copula, VAR, IoT sensor, daily active users and TCM in multivariate mode use sequential generation routines: they ignore the worker count and follow a fixed draw order rather than the per-series scheme described below. The design requirement is that changing the worker count or scheduling order must not change the generated output.

For the shared parallel path, series lengths and seeds are allocated before dispatch. For a request of $n$ series, a master random stream initialized with the user's seed draws, in order, $n$ series lengths, $n$ generation seeds and three perturbation seeds per series, corresponding to changepoints, anomalies and missingness. All three perturbation seeds are allocated even when their overlays are disabled. Each series is then generated using its assigned streams, independently of which worker executes the task.

Formally, series $i$ is
\begin{equation}
  y_i = \left(M_{m_i} \circ A_{a_i} \circ C_{c_i}\right)\left(G_\theta(g_i, n_i)\right),
  \label{eq:overlay}
\end{equation}
where $G_\theta$ denotes the generating mechanism with configuration $\theta$, $n_i$ is the series length, and $g_i$ is its generation seed. The configured changepoint, anomaly and missingness operators are denoted by $C$, $A$ and $M$, with seeds $c_i$, $a_i$ and $m_i$, respectively. A disabled overlay is the identity. Because seed allocation does not depend on which overlays are enabled, enabling an overlay changes the corresponding operator while preserving the base trajectory and the remaining seed assignments.

Hence, for a newly initialized generator, the output is determined by the seed, $n$ and the configuration within a fixed library version, execution path and numerical environment, and a subsequent call on the same instance continues the random stream. Two properties follow: output is invariant to the worker count, which the test suite checks by comparing one worker with four, and toggling a perturbation preserves the base trajectory and the seed assignments of the other overlays (Section~\ref{sec:overlays}).

Reproducibility also depends on the numerical environment. KernelSynth factorizes a covariance matrix using multithreaded linear algebra, so its output can change slightly with the BLAS thread count. We have not established a general bound across machines or numerical backends. The threaded and joint generators also use \texttt{NumPy}'s random generator, whose streams are not guaranteed to remain unchanged across releases. Reproducing a corpus therefore requires recording the \texttt{SynForecast} and \texttt{NumPy} versions, numerical backend and thread configuration, alongside a hash identifying the generated corpus.

The test suite checks both properties (\ref{app:overlaytest}). Because cross-version output stability is not guaranteed, benchmarks intended to remain comparable must pin both \texttt{SynForecast} and \texttt{NumPy}; our experiments use \texttt{SynForecast} 0.2.0, and the hourly frequency follow-up of Section~\ref{sec:hourly} uses a later commit.

The convenience pools seed each instance with the base seed plus a fixed offset, which reaches about 2{,}000 in \texttt{pretraining\_pool}. To avoid reusing instance seeds across replicates, base seeds must therefore be separated by more than the largest offset, as in our experiments.

\subsection{Augmentation}
\label{sec:augmentation}

The generators described so far sample unconditionally. Augmentation generates synthetic series conditional on observed data. For each observed series, \texttt{SynForecast} selects and fits one of nine generators with parameter-fitting support, then emits synthetic series identified as \texttt{\{original\_id\}\_aug\_\{i\}}, keeping them distinguishable from the originals. Three non-parametric alternatives are also available: a moving block bootstrap of the remainder of a classical decomposition, adapted from the Box--Cox and STL procedure of \citet{bergmeir2016bagging}, averaging under dynamic time warping \citep{petitjean2011global}, and mixup across series following TSMixup \citep{ansari2024chronos}. When fitting fails, the default behavior is to raise an error. An opt-in fallback generates a matched AR(1) series and reports every substitution, so a failed fit cannot silently change the generating mechanism.

Conditioning on observed data introduces a leakage risk. If augmentation uses observations from the evaluation window, information from that window can enter the training set. An interface can help enforce or audit this boundary by separating fitting from sampling, accepting an explicit cutoff or recording provenance, but each approach still depends on the caller specifying the held-out data correctly. \texttt{SynForecast} keeps synthetic series identifiable, but an identifier cannot show that a series was generated from training data alone. Hence, the caller must separate training and evaluation data before augmentation. We leave the evaluation of augmentation to future work.

\subsection{Throughput}
\label{sec:throughput}

We treat generation speed as a design consideration, since \citet{oreshkin2026sarsim} have already demonstrated on-the-fly generation. Generation is implemented largely in Rust and exposed through \texttt{PyO3}. On a machine with 20 cores, the 54-instance pool peaks at 419 series per second with eight workers, 2.45 times the throughput of one worker. Cost is dominated by series length and by KernelSynth: a daily pretraining draw, with series of up to 9,919 observations, takes 391~seconds against 1.3~seconds for quarterly (\ref{app:cost}).

\section{Results}
\label{sec:results}

In this section, we first describe the experimental protocol. Next, we compare the learned models with statistical baselines, evaluate fine-tuning, and study how the benefit of the prior depends on history length. We then analyze the failure of the prior at the hourly frequency and ablate the mechanism families. We end the section with an experiment that uses the overlays to measure the cost of structural breaks.

\subsection{Protocol}
\label{sec:protocol}

We pretrain small models on synthetic data produced by \texttt{SynForecast} and evaluate them on held-out real observations. No real data enters pretraining, and no evaluation observation enters any fitting step. The decision rules, test families and effect measures were specified before the experiments ran. \ref{app:prespec} compares each expectation with its outcome and documents all deviations from the plan, the most consequential of which was running the ablation on 20 replicates rather than the planned 10.

\emph{Data.} We use M4 \citep{makridakis2020m4} at the hourly, daily, weekly, monthly and quarterly frequencies. Yearly data are excluded because they lack the seasonal period required by the analysis in Section~\ref{sec:crossover}. Each cell evaluates a subsample of 300 series selected using a digest of the frequency and replicate, so all cells within a given frequency and replicate evaluate the same series. Weekly and hourly contain only 359 and 414 series, respectively, so replicates overlap in most of their series. We use natural seasonal periods (hourly 24, daily 7, weekly 52, monthly 12, quarterly 4). Our seasonal MASE values at daily and weekly frequencies are therefore not directly comparable to published M4 results, which treat these frequencies as non-seasonal.

\emph{Pretraining corpus.} We use the \texttt{pretraining\_pool} of \texttt{SynForecast} 0.2.0, comprising 54 generator instances from 19 classes. Every periodic instance is set to the target frequency's seasonal period, which the harness verifies for every corpus. Each (frequency, replicate) corpus is generated once and shared by both models. We draw 26 series per instance; the control keeps the first 20 of each and discards those containing non-finite values or too few observations for a training window. One IoT-sensor instance produces non-finite values in almost every series, so 53 instances contribute to all but 7 of the 150 corpora, in which it keeps a single series. The three mixture-autoregressive instances are non-seasonal, and the synthetic length ranges follow those of the M4 training series at each frequency.

\emph{Models.} NHITS \citep{challu2023nhits} and LightGBM \citep{ke2017lightgbm}, which we run via \texttt{NeuralForecast} and \texttt{MLForecast}, respectively, represent different function classes: a neural sequence model and a tree ensemble over lag features. They share the corpus, per-series standardization, input window and evaluation series, so agreement between them provides replication across function classes within a common pipeline. Both models use an input window equal to the forecast horizon: 48 observations for hourly data, 14 for daily, 13 for weekly, 18 for monthly and 8 for quarterly, with NHITS using this as its input size and LightGBM using lags 1 to $h$. This window is short by the standards of deep models on M4 and, at weekly, shorter than one seasonal cycle.

\emph{Conditions.} At each history level we evaluate \texttt{from\_scratch}, trained on the target series for 300 steps; \texttt{zero\_shot}, pretrained for 1,000 steps on synthetic data only; \texttt{pretrain\_ft}, the pretrained model trained for a further 300 steps on the target at a tenth of the learning rate; and Seasonal Naive and AutoETS \citep{hyndman2002state}, computed on the same split. The learned conditions train for fixed budgets, with no validation split, early stopping or tuning of the budget. Every fit records its realized number of training steps, and the harness rejects an NHITS fit that deviates from its budget. On ten replicates NHITS also reports \texttt{from\_scratch\_long}, trained for 1,000 steps (Section~\ref{sec:crossover}). LightGBM reports \texttt{from\_scratch} and \texttt{zero\_shot} only, because continuing to add trees to a booster fitted on synthetic data is not a comparable analog of fine-tuning.

\emph{Metric and comparisons.} Accuracy is the mean absolute scaled error \citep{hyndman2006another} of each series, in the seasonal form used in M4 \citep{makridakis2020m4}, which scales by the in-sample error of the seasonal naive forecast,
\begin{equation}
  \mathrm{MASE} = \frac{\frac{1}{h}\sum_{k=1}^{h} \left| y_{T+k} - \hat{y}_{T+k} \right|}{\frac{1}{T-m}\sum_{t=m+1}^{T} \left| y_t - y_{t-m} \right|},
  \label{eq:mase}
\end{equation}
in which $T$ denotes the training length, $h$ the horizon and $m$ the seasonal period. Conditions are compared within a replicate $r$ through the relative improvement of condition $b$ over condition $a$,
\begin{equation}
  \Delta_{a \to b}(r) = 100 \times \frac{\overline{\mathrm{MASE}}_a(r) - \overline{\mathrm{MASE}}_b(r)}{\overline{\mathrm{MASE}}_a(r)},
  \label{eq:improvement}
\end{equation}
where $\overline{\mathrm{MASE}}_c(r)$ denotes the mean MASE across the series in a cell for condition $c$ and replicate $r$, with a positive $\Delta$ indicating that condition $b$ outperforms condition $a$. Inference is across replicates, at each history level separately, with Wilcoxon signed-rank tests \citep{wilcoxon1945} and Holm correction \citep{holm1979}.

We run 30 replicates, each with its own series subsample, corpus and initializations; the ablation arms run on the first 20. \ref{app:protocol} gives the remaining details.

\subsection{Comparison with statistical baselines}
\label{sec:baselines}

We begin with the comparison that bounds every other result in this section. Table~\ref{tab:baselines} reports mean seasonal MASE of the statistical baselines and the best learned condition at the shortest and the longest history of each frequency.

\begin{widetable}[tbp]
\centering
\caption{Mean seasonal MASE, control arm, over 30 replicates and 300 series, with the standard error across replicates in parentheses. ``Best learned'' is the lowest of \texttt{from\_scratch}, \texttt{zero\_shot} and \texttt{pretrain\_ft} for NHITS, or of the first two for LightGBM. Lower is better; the best value in each row is bold.}
\label{tab:baselines}
\small\setlength{\tabcolsep}{3pt}
\begin{tabular}{llrrrr}
\toprule
& & \multicolumn{2}{c}{Statistical} & \multicolumn{2}{c}{Best learned} \\
\cmidrule(lr){3-4}\cmidrule(lr){5-6}
Frequency & History & Seas.\ Naive & AutoETS & NHITS & LightGBM \\
\midrule
Hourly & 96 obs & \textbf{1.293\,(0.004)} & 1.839\,(0.006) & 2.065\,(0.021) & 1.743\,(0.010) \\
 & full & \textbf{1.195\,(0.004)} & 1.603\,(0.006) & 1.412\,(0.012) & 1.443\,(0.006) \\
Daily & 28 obs & 1.490\,(0.013) & \textbf{1.229\,(0.013)} & 1.418\,(0.013) & 1.390\,(0.012) \\
 & full & 1.441\,(0.015) & 1.205\,(0.037) & \textbf{1.150\,(0.014)} & 1.332\,(0.065) \\
Weekly & 104 obs & 1.151\,(0.002) & \textbf{0.723\,(0.004)} & 0.777\,(0.004) & 0.793\,(0.005) \\
 & full & 1.218\,(0.004) & \textbf{0.572\,(0.002)} & 0.581\,(0.003) & 0.616\,(0.002) \\
Monthly & 36 obs & 1.393\,(0.011) & \textbf{1.152\,(0.010)} & 1.496\,(0.014) & 1.410\,(0.013) \\
 & full & 1.260\,(0.012) & \textbf{0.953\,(0.010)} & 1.178\,(0.013) & 1.149\,(0.013) \\
Quarterly & 16 obs & 1.522\,(0.013) & \textbf{1.243\,(0.014)} & 1.710\,(0.013) & 1.695\,(0.013) \\
 & full & 1.595\,(0.013) & \textbf{1.161\,(0.010)} & 1.552\,(0.012) & 1.617\,(0.015) \\
\bottomrule
\end{tabular}
\end{widetable}

We find that a statistical method attains the lowest error in nine of the ten rows: AutoETS in seven and Seasonal Naive in two. The single exception, daily at full history, is won by NHITS trained from scratch by 0.055 MASE with a paired standard error of 0.032, which is not a clear win. At the hourly frequency, Seasonal Naive performs best at both history levels, and the NHITS zero-shot model has 2.4 to 5.4 times its MASE. However, NHITS trained from scratch for 1,000 steps instead of 300 approximately matches Seasonal Naive at full history (1.192 against 1.188 on the ten replicates that ran it), which suggests that the gap between training from scratch and Seasonal Naive is largely a training-budget effect. Hence, at the scale studied here (a neural network of 156K to 191K parameters and a gradient-boosted ensemble), a practitioner should prefer none of the learned conditions. Where we report below that a synthetic prior beats training from scratch, the claim compares two learned conditions and is not a recommendation to use either.

Two aspects of the comparison may favor the statistical baselines. AutoETS selects its model form from the data, while the learned models run at a fixed budget without tuning, and the synthetic M4 timestamps offer no calendar information that a neural model could exploit. Hence, the margins in Table~\ref{tab:baselines} reflect the evaluated configurations and do not isolate the effect of method class. At monthly and quarterly frequencies, AutoETS leads by 0.20 to 0.45 MASE with paired standard errors of at most 0.015, so the gaps are large relative to their sampling uncertainty. The smaller gaps at weekly and daily frequencies may be more sensitive to tuning.

\subsection{Fine-tuning after synthetic pretraining}
\label{sec:null}

We examine whether fine-tuning a pretrained model improves forecasting accuracy compared with training from scratch on the target data or using the pretrained model without adaptation. Figure~\ref{fig:conditions} shows both comparisons for each frequency and history level, alongside the comparison between the pretrained model and a model trained from scratch examined in Section~\ref{sec:crossover}. Table~S4 in the supplementary material reports the corresponding statistical tests for the fine-tuning comparisons.

\begin{widefigure}[tbp]
\centering
\includegraphics[width=\linewidth]{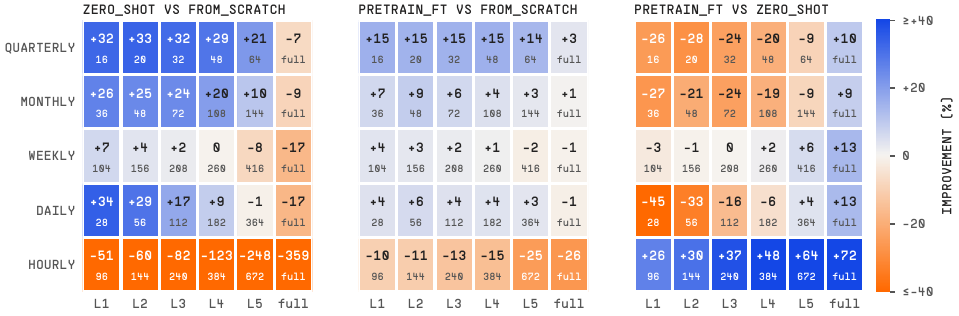}
\caption{Relative improvement $\Delta_{a\to b}$ (Eq.~\ref{eq:improvement}, \%) of the condition named first over the condition named second, NHITS, mean over 30 replicates, by frequency and history level; the small number in each cell is the history in observations; Table~S2 in the supplementary material gives each value with its 95\% interval. Bold values are significant after Holm correction within the 30 tests of a panel (Wilcoxon signed-rank test across replicates). Blue favors the first condition, orange the second; the color scale is clipped at $\pm 40$\%.}
\label{fig:conditions}
\end{widefigure}

\emph{Fine-tuning against training from scratch.} We find that fine-tuning reduces forecast error for at least some history levels at four of the five frequencies. The largest average reduction occurs at quarterly, at approximately 13\%, with statistically significant differences at all six history levels. At monthly, fine-tuning reduces error at every level. At daily and weekly, the improvement broadly diminishes as history length increases and turns negative at the longest histories. However, at the hourly frequency, fine-tuning increases error by 10 to 26\% across all history levels. With daily cycles in the corpus, it reduces error at every level (Section~\ref{sec:hourly}). Overall, 28 of the 30 comparisons are statistically significant after correction for multiple testing.

\emph{Fine-tuning against the pretrained model.} Compared with the pretrained model without adaptation, fine-tuning increases error by 26 to 45\% at the shortest history for quarterly, monthly and daily data. At full history, it reduces error by 9 to 13\% at all frequencies except hourly. These results indicate that a fixed budget of 300 fine-tuning steps, without early stopping, can substantially reduce the benefits of synthetic pretraining when only a few dozen observations per series are available. Selecting the budget on held-out validation data, with zero steps among the candidates, could mitigate this effect, although we did not evaluate this approach. At the hourly frequency, fine-tuning reduces error relative to the zero-shot model at every history level, which reflects the poor zero-shot performance discussed in Section~\ref{sec:hourly}.

Hence, no learned configuration dominates across frequencies and history lengths. At short histories, the pretrained model without adaptation performs best at the non-hourly frequencies. At full history, fine-tuning and training from scratch are equivalent within a 5\% margin at these four frequencies (two one-sided tests, $p < 0.02$), although equivalence within a 2\% margin is not established at daily and quarterly. The direction of the fine-tuning effect is consistent across replicate subsets. When the replicates are split into two groups of 15, the effect has the same sign in both groups for all 30 comparisons. We pre-specified that the quarterly improvement would be treated as a robust result only if fine-tuning were also not significantly worse than the zero-shot model at any history level. This criterion is not met because fine-tuning is worse at short histories.

\ref{app:protocol} describes how the statistical tests are constructed.

\subsection{Effect of history length}
\label{sec:crossover}

We find that the benefit of the synthetic prior depends on history length (Figure~\ref{fig:conditions}, left). Compared with training from scratch, the prior reduces forecast error by 26 to 34\% at the shortest histories for daily, monthly and quarterly data, with improvements in all 30 replicates, but increases error by 7 to 17\% at full history. The direction of the effect reverses at four of the five frequencies for both models, although the initial advantage of the prior is negligible for LightGBM at weekly ($+0.02\%$). At the hourly frequency, the prior increases error at every history level. Its error is 51\% higher at 96 observations and 4.6 times that of training from scratch at full history. Section~\ref{sec:hourly} traces this failure to the corpus, which lacks daily cycles. With these cycles, the NHITS prior changes sign at the hourly frequency as well.

We examine whether the crossover aligns across frequencies under different measures of history length. Within a frequency, observation count and seasonal cycles differ only by a constant factor; comparing frequencies allows these measures to be distinguished. For each frequency $f$ with an observed reversal, we locate the crossover $c_f$, defined as the realized history length at which $\Delta_{\texttt{from\_scratch}\to\texttt{zero\_shot}}$ crosses zero. We express it in raw observations $c_f$, seasonal cycles $c_f/m_f$, and history relative to the forecast horizon $c_f/h_f$, where $m_f$ and $h_f$ denote the seasonal period and forecast horizon, respectively.

Let $\mathrm{CV}_k$ denote the coefficient of variation of normalization $k$ across frequencies with an observed crossover, and let $\mathrm{CV}_{(1)} \leq \mathrm{CV}_{(2)}$ denote the two smallest values. Under our pre-specified rule, the normalization with the smallest coefficient of variation is considered to govern the crossover if
\begin{equation}
  \mathrm{CV}_{(1)} < 0.25 \quad\text{and}\quad \mathrm{CV}_{(2)} \geq 2\,\mathrm{CV}_{(1)}.
  \label{eq:rule}
\end{equation}
The rule and the interpretation of each possible outcome were specified before the experiments, using the nominal history axis. \ref{app:prespec} reports that analysis, which yields the same conclusion as the analysis based on realized history.

\begin{widefigure}[tbp]
\centering
\includegraphics[width=\linewidth]{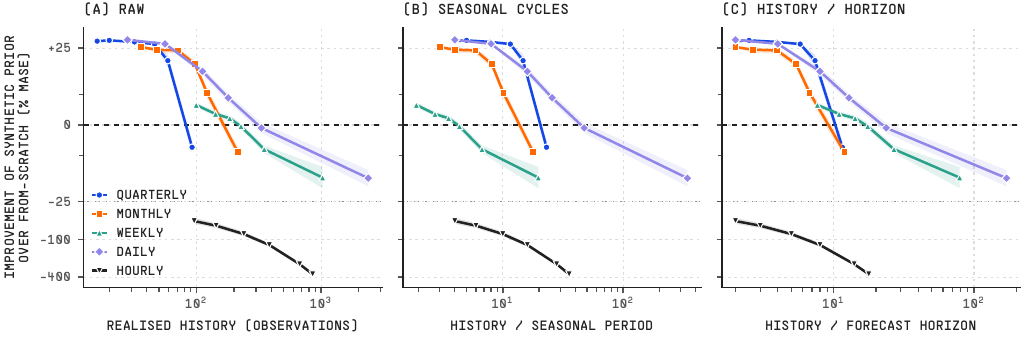}
\caption{Improvement of the synthetic prior (\texttt{zero\_shot}) over target-specific training (\texttt{from\_scratch}) against history, NHITS, under the three candidate normalizations. Bands are 95\% intervals over 30 replicates; the dashed line is parity. The vertical scale is linear within $\pm25$\% and logarithmic beyond it, with the transition marked by a dotted rule: hourly reaches $-359$\%, which on a linear scale compresses the four frequencies that carry the argument into an illegible band around parity. No point is clipped. All three panels are shown because none of them collapses the five frequencies onto a common curve: in (b), seasonal cycles, the crossings span 4.2 (weekly) to 45.3 (daily). Hourly is negative throughout.}
\label{fig:crossover}
\end{widefigure}

\begin{widetable}[tbp]
\centering
\caption{Crossover location by frequency and normalization, NHITS, in realized training observations, over 30 replicates. Hourly is excluded because the effect is negative at every level; that exclusion was the pre-specified handling for a frequency with no crossing. CV is the coefficient of variation across the four remaining frequencies, with a bootstrap 95\% interval over replicates. Figure~\ref{fig:crossover} is generated independently from the same cells and its interpolated crossings agree with this table to within 0.1 observations.}
\label{tab:crossover}
\begin{tabular}{lrrr}
\toprule
Frequency & Raw (obs) & Cycles & History/horizon \\
\midrule
Quarterly & 83.7 & 20.9 & 10.5 \\
Monthly   & 171.9 & 14.3 & 9.5 \\
Weekly    & 220.1 & 4.2 & 16.9 \\
Daily     & 316.8 & 45.3 & 22.6 \\
\midrule
CV        & 0.491 & 0.824 & 0.411 \\
95\% CI   & [0.459, 0.628] & [0.783, 0.975] & [0.375, 0.556] \\
\bottomrule
\end{tabular}
\end{widetable}

We find that none of the three normalizations satisfies the pre-specified crossover criterion (Figure~\ref{fig:crossover}, Table~\ref{tab:crossover}). History relative to the horizon has the lowest dispersion, with coefficients of variation of 0.411 for NHITS and 0.371 for LightGBM, both above the threshold of 0.25. The corresponding ratios $\mathrm{CV}_{(2)}/\mathrm{CV}_{(1)}$ are 1.19 and 1.27, also below the required value of 2. Seasonal cycles produce the greatest dispersion, contrary to the original expectation motivated by the need for repeated cycles to identify seasonal patterns. The crossover occurs at 4.2 cycles for weekly data and 45.3 cycles for daily data. This ranking depends partly on weekly, where the 13-observation input window covers only a quarter of the 52-week cycle, so neither model has access to a seasonal lag. Excluding weekly yields similar dispersion for seasonal cycles and raw observations (NHITS CV 0.61 and 0.62, respectively). Because the forecast horizon equals the input-window length, normalization by the horizon is also normalization by the receptive field, so this comparison cannot distinguish their roles.

However, the crossover could also reflect an under-trained comparator under the 300-step budget. On the ten replicates that additionally trained from scratch for 1,000 steps (Table~S5 in the supplementary material), the longer budget reduces error by 16 to 23\% at every hourly history level from 144 observations onward. At the other frequencies, it increases error by up to 5.3\% at most shorter histories, while differences at full history remain within 2.2\%. The longer budget therefore changes the balance between short- and long-history performance, shifting the crossovers to 83, 169, 236 and 328 observations without changing the conclusion of the normalization analysis. For the test used, ten replicates do not permit statistical significance after Holm correction across 30 comparisons, so we report effect estimates and their intervals.

Under the main protocol, the observed crossovers fall between 84 and 317 realized observations for NHITS and between 66 and 263 for LightGBM. This is broadly in line with ForecastPFN, which obtained the most MSE wins against other model classes at training budgets of 50--250 observations \citep{dooley2023forecastpfn}, although that comparison used different comparators, metrics and datasets. The comparator is the same network trained from scratch at a fixed budget and, except at weekly, performs worse than Seasonal Naive at the shortest histories. At quarterly, monthly and daily, the NHITS prior's error relative to Seasonal Naive remains approximately constant with history length, whereas at weekly it falls from 0.67 to 0.56 times that of Seasonal Naive (Table~S6 in the supplementary material). The crossover occurs where the comparator's learning curve intersects this level, and improved regularization of the comparator could shift its location. At the non-hourly frequencies, the synthetic prior generally performs better with a few dozen observations per series, while training from scratch performs better with several hundred. AutoETS nevertheless outperforms both at this scale (Section~\ref{sec:baselines}).

\subsection{Failure of the prior at the hourly frequency}
\label{sec:hourly}

In the main experiment, hourly is the one frequency at which the synthetic prior never helps, and the failure grows with history: the zero-shot error relative to Seasonal Naive rises from 2.4 at 96 observations to 5.4 at full history, although the model only sees the last 48 observations. We tested three explanations. We tested the third, periodicity, in a follow-up run after the main experiment.

\emph{Normalization.} NHITS inputs are standardized per series over the whole retained history, so a longer history rescales the 48 observations the model sees. When we rerun the hourly grid of the main experiment with each input window standardized instead, pretraining included, the ratio to Seasonal Naive no longer grows (2.1 at 96 observations, 1.9 at full history), and zero-shot error falls by 11 to 65\%, significantly at every level. We ran this for NHITS only. LightGBM, which standardizes the same way, shows the same growth (from 3.1 to 6.4 times Seasonal Naive), but we did not rerun it.

\emph{Coverage.} We placed each frequency's evaluated series, the pretraining corpus and an equally sized sample of other M4 series in one feature space fitted on the evaluated series, using \texttt{SynForecast}'s coverage module, and counted the evaluated series in grid cells the comparison sample does not reach (Table~\ref{tab:coverage}). A size-matched sample of the corpus leaves 72\% of the hourly series uncovered, against 24\% for real hourly series. At the other frequencies, the gap is 7 to 26 points, and hourly has the largest gap at grid resolutions of 20, 30 and 40 bins per axis. However, the absolute figures are noisy and depend on sample size. Across replicates, the hourly fraction ranges from 0.60 to 0.85, and the full corpus of about 1{,}060 series, instead of the matched 114, leaves 47\% uncovered. Measured instead by the share of all grid cells that real series occupy and the corpus misses, the hourly gap is not the largest.

\begin{widetable}[tbp]
\centering
\caption{Feature-space coverage of the evaluated M4 series, mean over replicates 0--2. ``Uncovered'' is the fraction of evaluated series lying in grid cells (30 $\times$ 30, PCA fitted on the evaluated series) that the comparison sample does not occupy. The comparison samples are the round's pretraining corpus and other M4 series of the same frequency, subsampled to the same size, $n$; at weekly and hourly $n$ is limited by the series M4 has outside the evaluated subsample. ``Full'' scores the whole control corpus instead of a sample of size $n$. At weekly, 54 to 56 of the 300 evaluated series have undefined features and are dropped.}
\label{tab:coverage}
\begin{tabular}{lrrrrr}
\toprule
& & \multicolumn{3}{c}{Uncovered evaluated series} & \\
\cmidrule(lr){3-5}
Frequency & $n$ & Synthetic & Real & Gap & Full \\
\midrule
Quarterly & 1000 & 0.27 & 0.10 & $+0.17$ & 0.26 \\
Monthly   & 1000 & 0.38 & 0.11 & $+0.26$ & 0.35 \\
Weekly    & 59 & 0.44 & 0.29 & $+0.14$ & 0.09 \\
Daily     & 1000 & 0.10 & 0.03 & $+0.07$ & 0.10 \\
Hourly    & 114 & 0.72 & 0.24 & $+0.48$ & 0.47 \\
\bottomrule
\end{tabular}
\end{widetable}

Normalization explains why the failure grows with history, but not the failure itself: with window standardization, the prior is still worse than training from scratch at every hourly level, and about twice as bad as Seasonal Naive.

\emph{Periodicity.} Real hourly series are dominated by a daily cycle. Every periodic instance in the hourly corpus has period 24, but few instances are periodic. Measured as the strength of the 24-step seasonality, $1 - \mathrm{Var}(\text{remainder})/\mathrm{Var}(\text{detrended})$ under a classical decomposition, the median over the series of a corpus is 0.03 on average across replicates, against 0.92 for the evaluated M4 series. We therefore reran the hourly grid on a second corpus that adds calendar cycles to the control corpus. \texttt{SynForecast}'s \texttt{calendar\_overlays} prior gives each pool instance, with probability 0.8, a time-of-day profile of one to five harmonics and, with probability 0.7, a day-of-week profile. The profile is either added, with a peak of one to eight standard deviations, or, on strictly positive series, multiplied in, with a swing of 20 to 90\% of the level. Each instance keeps the series of the control corpus, so the two corpora differ only in the cycles, and the median seasonal strength rises to 0.60. Note that no model receives a calendar feature, so a model can learn that 24- and 168-step cycles occur, but not their phase. The follow-up uses all 30 replicates and six history levels, NHITS with both scalers, and LightGBM. Training from scratch and the baselines are shared with the control corpus of the same run, whose zero-shot error relative to Seasonal Naive is within 3\% of the main experiment at every level.

We find that calendar cycles reduce zero-shot error by 43 to 62\% for NHITS and by 52 to 63\% for LightGBM, at every history level and in every replicate (Table~\ref{tab:calendar}). With cycles, the NHITS prior outperforms training from scratch by 16\% at 96 observations and by 12\% at 144, and loses by 70\% at full history. The sign changes at 323 realized observations, just above the range of 84 to 317 observations at the other frequencies. In addition, fine-tuning the prior with cycles outperforms training from scratch at every level, by 9 to 21\%. However, the prior with cycles remains 1.3 to 2.5 times worse than Seasonal Naive. For LightGBM, and for NHITS with the window scaler, it also still loses to training from scratch at every level, by 10 to 106\% and by 9 to 24\%, respectively.

\begin{table}[tbp]
\centering
\caption{Hourly follow-up: the pretraining corpus without (control) and with calendar cycles, mean over 30 replicates. Top: zero-shot MASE relative to Seasonal Naive. Bottom: improvement $\Delta$ (Eq.~\ref{eq:improvement}, \%) of the zero-shot model over the same model trained from scratch; bold values are significant after Holm correction across the six history levels (Wilcoxon signed-rank test across replicates). Series and window name NHITS's input standardization. The control rows, including those with the window scaler, come from the follow-up run, not from the main experiment, so they differ slightly from the main-experiment values quoted in the text.}
\label{tab:calendar}
\footnotesize
\setlength{\tabcolsep}{3pt}
\begin{tabular}{@{}lrrrrrr@{}}
\toprule
& \multicolumn{6}{c}{History (observations)} \\
\cmidrule(l){2-7}
& 96 & 144 & 240 & 384 & 672 & Full \\
\midrule
\multicolumn{7}{@{}l}{\emph{Zero-shot MASE relative to Seasonal Naive}} \\
NHITS, series & 2.34 & 2.44 & 2.72 & 3.28 & 4.52 & 5.38 \\
\quad with cycles & 1.34 & 1.37 & 1.44 & 1.56 & 1.80 & 2.01 \\
NHITS, window & 2.17 & 2.11 & 2.07 & 2.02 & 1.98 & 1.91 \\
\quad with cycles & 1.51 & 1.48 & 1.46 & 1.44 & 1.42 & 1.39 \\
LightGBM & 3.17 & 3.19 & 3.46 & 4.25 & 5.83 & 6.37 \\
\quad with cycles & 1.49 & 1.51 & 1.61 & 1.79 & 2.15 & 2.49 \\
\midrule
\multicolumn{7}{@{}l}{\emph{$\Delta_{\texttt{from\_scratch}\to\texttt{zero\_shot}}$ (\%)}} \\
NHITS, series & \textbf{$-$47} & \textbf{$-$57} & \textbf{$-$79} & \textbf{$-$119} & \textbf{$-$245} & \textbf{$-$357} \\
\quad with cycles & \textbf{$+$16} & \textbf{$+$12} & \textbf{$+$5} & $-$4 & \textbf{$-$37} & \textbf{$-$70} \\
NHITS, window & \textbf{$-$65} & \textbf{$-$54} & \textbf{$-$75} & \textbf{$-$70} & \textbf{$-$70} & \textbf{$-$59} \\
\quad with cycles & \textbf{$-$15} & \textbf{$-$9} & \textbf{$-$24} & \textbf{$-$22} & \textbf{$-$23} & \textbf{$-$16} \\
LightGBM & \textbf{$-$135} & \textbf{$-$146} & \textbf{$-$162} & \textbf{$-$196} & \textbf{$-$312} & \textbf{$-$428} \\
\quad with cycles & \textbf{$-$10} & \textbf{$-$17} & \textbf{$-$22} & \textbf{$-$25} & \textbf{$-$52} & \textbf{$-$106} \\
\bottomrule
\end{tabular}
\end{table}

To conclude, the lack of daily cycles in the corpus explains most of the hourly failure, and normalization explains its growth with history. With cycles, the NHITS ratio to Seasonal Naive still rises 1.5-fold from 96 observations to full history with the series scaler, and does not grow with the window scaler. The coverage results are consistent with this, as hourly is the frequency the corpus covers worst. However, the coverage comparison is descriptive, its hourly gap of 0.48 falls just below the pre-specified threshold of 0.50, and we did not recompute it for the corpus with cycles. We specified the follow-up after the main experiment and after a five-replicate pilot (\ref{app:prespec}), and ran it on a commit of the library that is not yet released, without the generic overlays that the library's pretraining prior can add.

\subsection{Ablation of mechanism families}
\label{sec:ablation}

In this section, we use the capability that distinguishes mechanism specification (Section~\ref{sec:gratis}): removing a named family. We define seven families by the instances in the pretraining pool, remove each family in turn while keeping the corpus size approximately constant, and measure the change in zero-shot accuracy. Every arm keeps a prefix of the same per-instance draws as the control, and draws more from each remaining instance to keep the corpus size. Hence, an arm differs from the control both in the family removed and in the weight of the remaining families, and an effect measures a family against a reweighted remainder.

\begin{figure}[tbp]
\centering
\includegraphics[width=\linewidth]{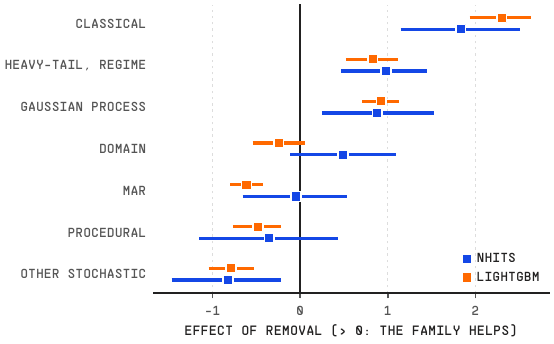}
\caption{Effect of removing each mechanism family on zero-shot accuracy (Eq.~\ref{eq:ablation}), primary population (daily, weekly, monthly and quarterly; three history levels each; 20 replicates), with 95\% intervals from a bootstrap over replicates. Positive: the family helps. Families: classical = SARIMA, ETS, INAR; Gaussian process = four GP instances; heavy-tail and regime = GARCH, L\'evy, regime switching; other stochastic = fractional Brownian motion, chaotic, bounded, cyclic; domain = intermittent demand, energy load, IoT sensor, vital signs; procedural = TSI, TCM, KernelSynth; MAR = three non-seasonal mixture-autoregressive instances. Table~\ref{tab:ablation} also reports the pre-specified replicates 0--9.}
\label{fig:ablation}
\end{figure}

The effect of removing family $a$ is
\begin{equation}
  E_a = \frac{100}{|\mathcal{C}|}\sum_{(f, l, r) \in \mathcal{C}} \frac{1}{N}\sum_{i=1}^{N} \left( \log \mathrm{MASE}^{(a)}_{i} - \log \mathrm{MASE}^{(0)}_{i} \right),
  \label{eq:ablation}
\end{equation}
in which $\mathcal{C}$ denotes the cells (frequency $f$, history level $l$, replicate $r$) of a population and $N$ the series of a cell, so that $E_a$ is positive when removing the family hurts. Intervals come from a bootstrap over replicates. The primary population excludes hourly, where the zero-shot model fails, and an effect is said to replicate when both models' intervals exclude zero with the same sign (Figure~\ref{fig:ablation}; Table~\ref{tab:ablation} in \ref{app:results}).

We find that one family replicates on the pre-specified replicates and four on the run as executed, and that all effects are small. We had pre-specified the ablation on replicates 0--9 and ran it on 0--19, a change decided before launch because ten replicate clusters give under-covering intervals (\ref{app:prespec}); Table~\ref{tab:ablation} reports both. Removing the classical family costs about 2\% for both models on both replicate sets. Removing the Gaussian-process or the heavy-tail and regime family costs about 1\%, and removing the remaining stochastic mechanisms -- fractional Brownian motion, chaotic systems, bounded processes and cycles -- improves accuracy by about 1\%. These three effects are supported by both models in the analysis using 20 replicates. On the pre-specified subset of 10, the LightGBM intervals exclude zero but the NHITS intervals do not, although the point estimates differ from those of the larger analysis by no more than 0.2 points. Each of the four effects keeps its sign at all four frequencies of the primary population for both models, with two exceptions for NHITS. The domain, procedural and MAR families do not replicate.

We highlight three observations. First, mechanism families are not interchangeable: removing the classical family changes downstream accuracy in a direction on which both model classes agree. Second, the effects are modest, at most 5\% at any non-hourly frequency, which suggests that corpus composition has a limited influence at this scale. By comparison, TempoPFN's leave-one-generator-out ablation, with a model of about 40M parameters, found that removing a single generator raised normalized CRPS on GIFT-Eval by 2--14\%, and by 26\% for its stochastic-differential-equation generator \citep{moroshan2025tempopfn}; the metric and the unit removed differ from ours. Third, the direction of an effect is not always what one would expect. The heavy-tail and regime family (GARCH and L\'evy processes, and the Markov-switching autoregression) contributes, whereas the long-memory, chaotic and bounded processes do not. The sign of the heavy-tail effect reversed between this run and our earlier run on a smaller pool, so we consider it provisional. The library makes such an ablation possible outside the training code of a single model, behind one interface. At the hourly frequency, the effects are larger, up to 19\%, and follow a different pattern, so we do not generalize from them.

\subsection{Effect of structural breaks}
\label{sec:overlaydemo}

\begin{figure}[tbp]
\centering
\includegraphics[width=\linewidth]{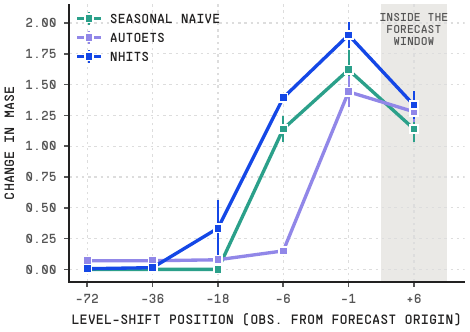}
\caption{Change in MASE caused by one level shift of three times the series' median standard deviation, by its position relative to the forecast origin; monthly SARIMA panels, 5 replicates of 300 series, mean $\pm$ one standard error across replicates. The shaded position lies inside the 18-step forecast window.}
\label{fig:overlay}
\end{figure}

Finally, we use the overlays of Section~\ref{sec:overlays} for a controlled stress test. We generate monthly SARIMA panels of 300 series of 144 observations and, from the same seeds, the same panels with a single level shift of three times the series' median standard deviation at a chosen position. Before the break, the perturbed series equal the unperturbed ones bit for bit, which the experiment asserts for every panel. Seasonal Naive, AutoETS and NHITS each forecast the last 18 observations of both versions, and Figure~\ref{fig:overlay} shows the paired change in MASE, measured on the unperturbed series' scale (values in Table~S3 of the supplementary material).

We find that the three methods fail in different ways, each traceable to how it uses history. Seasonal Naive is unaffected unless the break falls within its last season. AutoETS, which fits the whole history, pays a small cost for any break six or more observations back, and absorbs a break six observations back far better than the other methods. NHITS, which sees only the last 18 observations, is almost unaffected by breaks outside that window and strongly affected by breaks inside it. These results are as expected. Because the design holds everything else fixed, the comparison isolates how each method uses history.

\section{Discussion}
\label{sec:discussion}

\emph{Role of synthetic priors.} A synthetic prior provides a plausible starting point when target data are scarce, and its value decreases as more target data become available. Our results add two practical refinements. First, fine-tuning is not a free improvement on the prior. With a few dozen observations per series, a fixed 300-step fine-tune without early stopping removes much of what the prior provides, so in that regime the pretrained model is better used as is, although a classical baseline is better still. Second, the prior helps only when observations per series are few, and practitioners with long histories gain nothing from it. Note that our short histories are the recent tails of mature M4 series, and not the first observations of new products, whose launch ramps and intermittency M4 tails lack. Whether our results carry over to new products remains to be tested.

\emph{Diagnosis of a failing prior.} The hourly results show that a failing prior can have more than one cause, and that these causes can be separated (Section~\ref{sec:hourly}). Hence, before concluding that synthetic data does not suit a frequency, it is worth checking how inputs are normalized and whether the corpus contains the dominant structure of the target data. Seasonal strength and feature coverage are cheap checks, and a rerun that changes a single mechanism or overlay tests the diagnosis directly.

\emph{Relation to other findings.} \citet{cazaux2026synthetic} find that the effect of synthetic augmentation depends mainly on model architecture, with gains at reduced data limited to some combinations of architecture and dataset, whereas our results show that the effect of synthetic pretraining depends on the amount of available target history. The two findings are complementary, and together suggest that whether synthetic data helps has no general answer. Our results also differ from those of \citet{oreshkin2026sarsim} on GiftEval and of \citet{chronos2} on fev-bench, where synthetic or partly synthetic pretraining produces models that beat statistical baselines zero-shot. On the M-Series, which include M4, \citet{oreshkin2026sarsim} instead report mixed results against AutoARIMA, which is closer to what we observe. Moreover, our corpora are orders of magnitude smaller, our models are about 700 times smaller than Chronos-2 in parameter count (156K to 191K against 120M), and our input window is one horizon long, against contexts of up to 8,192 observations. Hence, we believe that synthetic pretraining may require a large scale, a long receptive field, or both, before it competes with fitting the target directly.

\emph{Limitations.} Limitations of our work are that we evaluate two model classes, trained on the same corpus and preprocessing, on one dataset family. Generalization across architectures and to other datasets, such as the non-M4 collections in the Monash archive \citep{godahewa2021monash}, is untested. Both learned conditions train for fixed budgets without validation or early stopping, the input window equals the forecast horizon, and no condition pretrains on real series, unlike \citet{kuvshinova2024synthesize}, who also pretrain on M3 and M4. Hence, our results compare a synthetic prior with one untuned way of training on the target, and cannot separate the value of synthetic data from that of pretraining in general. Finally, the authors develop the library, so this is not an independent evaluation. The declaration of competing interest at the end of this article states the relationship, and \ref{app:prespec} and the artifact allow the evaluation to be audited.

\emph{Future work.} For future work, we see four experiments that follow from the questions this paper leaves open:
\begin{enumerate}
  \item Varying corpus size while holding mechanism diversity fixed, and vice versa, would separate two explanations our results leave confounded: that the prior is too small, or that it has the wrong shape.
  \item Applying a large pretrained forecaster to the same splits would test whether synthetic pretraining escapes the crossover at scale, read as an upper bound on zero-shot accuracy rather than as a matched comparison.
  \item Comparing the moving block bootstrap, which preserves the observed trend and seasonality, with parametric augmentation, which replaces the process with a fitted approximation, on matched splits and panel sizes would show which of the two improves forecasts more.
  \item Rerunning the full grid on a corpus with calendar cycles, and adding mechanisms targeted at the regions of feature space that the corpus still does not reach, would show how much of the remaining gap to Seasonal Naive is due to the corpus rather than to the scale of the model.
\end{enumerate}

\section{Conclusion}
\label{sec:conclusion}

In this paper, we presented \texttt{SynForecast}, an open-source Python library for synthetic time series designed for forecasting research. Its 32 generators, 19 of which implement named mechanisms with a published source, share one interface. This interface makes it possible to ablate mechanism families outside the training code of a single model, and to stress-test forecasting methods on a shared base trajectory.

Using \texttt{SynForecast} on M4 with two small learned models, we find four results. First, at this scale, an automatically specified classical baseline attains the lowest error in nine of ten frequency-history combinations. Second, a synthetic prior outperforms the same network trained from scratch at a fixed budget by up to 34\% with a few dozen observations, and loses to it at full history at every frequency. At four frequencies, the sign changes between 84 and 317 observations, and none of the three rescalings of history we pre-specified predicts where. At the hourly frequency, where the prior fails, the corpus lacks daily cycles: adding them reduces zero-shot error by 43 to 63\% for both models under per-series standardization and gives the NHITS prior a crossover at 323 observations, while normalization explains why the remaining error grows with history. Third, fine-tuning the prior outperforms training from scratch at four frequencies, by about 13\% at quarterly, but at short histories a fixed 300-step fine-tune is worse than the prior used as is. Fourth, removing the classical family costs about 2\% for both models, and three further families have effects of about 1\% that replicate across models on 20 replicates, but not on the 10 we had pre-specified.

To conclude, the usefulness of synthetic data for forecasting depends on the setting, in ways that per-model studies have only begun to map. We believe that \texttt{SynForecast} provides a basis for extending these experiments to other architectures, other datasets, augmentation, and the corpus scales at which simulation-only training has recently succeeded.

\section*{CRediT author statement}

Olivier Sprangers: Conceptualization, Methodology, Software, Validation, Formal analysis, Investigation, Data curation, Visualization, Project administration, Writing -- original draft, Writing -- review and editing. Cristian Challu and Jan Rathfelder: Writing -- original draft, Writing -- review and editing. Han Wang: Software. Max Mergenthaler Canseco: Writing -- review and editing.

\section*{Funding}

This research did not receive any specific grant from funding agencies in the public, commercial, or not-for-profit sectors.

\section*{Declaration of competing interest}

The authors work at Nixtla, the company that develops the software examined as the principal subject of this article. This relationship may be perceived as a competing interest.

\section*{Data and code availability}

All datasets used are public and obtained programmatically; no new dataset is released and no source file is redistributed. \texttt{SynForecast} is released under the Apache 2.0 license at \url{https://github.com/Nixtla/synforecast} and on PyPI; every experiment ran against \texttt{synforecast 0.2.0} as published there, except the hourly follow-up of Section~\ref{sec:hourly}. That follow-up ran against commit \texttt{b2338c8} of the repository, which adds the calendar cycles and was not yet part of a release.

The accompanying artifact contains the experiment runners, which record the package versions, seeds and realized training steps of every cell; the analysis scripts; the pre-specification; the result rows of every cell; and a manifest mapping each table and figure to the script and inputs that produce it, so that every table and figure can be regenerated without a GPU. It is publicly available at \artifacturl.

\section*{Declaration of generative AI and AI-assisted technologies in the writing process}

During the preparation of this work the authors used OpenAI Codex and Anthropic Claude Code to assist with literature discovery and drafting. The authors reviewed and edited the content and take full responsibility for the content of the article.

\bibliographystyle{elsarticle-harv}
\bibliography{references}

\appendix

\section{Existing generators}
\label{app:generators}

Table~\ref{tab:generators} scores 11 existing generators against the four requirements of Section~\ref{sec:intro}. We included generators that produce series from a model rather than transforming observed ones, and that are either packaged for reuse or were used to pretrain a published forecasting model; we excluded learned generative models, which are fitted to data and are closer to augmentation. One included generator, \texttt{generate\_series} in \texttt{utilsforecast}, belongs to the same ecosystem as \texttt{SynForecast}. Every score comes from reading the released code and documentation at the stated version; nothing was executed. SarSim0 has no official code release, so only its design, which the paper describes, is scored. The requirements are \texttt{SynForecast}'s design goals (Section~\ref{sec:intro}), so the table positions \texttt{SynForecast} rather than ranking tools built for other purposes.

\begin{widetable}[tbp]
\centering
\footnotesize
\setlength{\tabcolsep}{4pt}
\caption{Existing generators against the requirements of Section~\ref{sec:intro}: (i) output carries a frequency, a time origin and a series identity; (ii) a perturbation can be varied while the base series stays bit-identical; (iii) output is identical under parallel execution; (iv) many named generating processes, each removable. \emph{Yes} means fully met. For (ii), \emph{partly} means perturbations act on a generated series but their isolation follows from draw order or a shared random stream and is not guaranteed. For (iii), \emph{partly} means the library exposes a seed but either generates in one process or gives no guarantee across worker counts; \emph{no} means no seed is exposed. For (iv), \emph{yes} means ten or more mechanisms the user can select individually, \emph{partly} two to nine, or primitives the user composes, and \emph{no} a single family whose components the user cannot select. A dash marks a property that cannot be assessed without code. \texttt{TimeSynth} has no publication; its repository, last updated in 2020, is \url{https://github.com/TimeSynth/TimeSynth}.}
\label{tab:generators}
\begin{tabular}{@{}llccccl@{}}
\toprule
Generator & Version inspected & (i) & (ii) & (iii) & (iv) & Base mechanisms \\
\midrule
\texttt{gratis} (R) \citep{kang2020gratis} & 1.0.8 & yes & no & partly & partly & 3: MAR, ARIMA, ETS \\
\texttt{GluonTS} \texttt{dataset.artificial} \citep{alexandrov2020gluonts} & commit \texttt{d60556c} & yes & partly & partly & partly & 2 presets, about 14 primitives \\
\texttt{Darts} generators \citep{herzen2022darts} & 0.47.0 & partly & no & no & partly & 6 \\
\texttt{TimeSynth} &  0.2.4 & no & partly & no & partly & 7 \\
\texttt{TSGM} simulators \citep{nikitin2023tsgm} & 0.1.0 & no & partly & partly & partly & 3 simulators \\
\texttt{GutenTAG} \citep{gutentag} & 1.5.0 & partly & partly & yes & yes & 13 base oscillations \\
KernelSynth \citep{ansari2024chronos} & commit \texttt{1224faf} & partly & no & no & no & 1: composed GP kernels \\
ForecastPFN prior \citep{dooley2023forecastpfn} & commit \texttt{919dda4} & partly & no & no & no & 1: trend $\times$ season $\times$ noise \\
\texttt{utilsforecast} & 0.2.16 & yes & no & partly & no & 1: sawtooth plus noise \\
TempoPFN pipeline \citep{moroshan2025tempopfn} & commit \texttt{5969ec6} & yes & partly & partly & yes & 14 \\
SarSim0 \citep{oreshkin2026sarsim} & no official code & -- & -- & -- & no & 1: SARIMA \\
\midrule
\texttt{SynForecast} & 0.2.0 & yes & yes & yes & yes & 32 \\
\bottomrule
\end{tabular}
\end{widetable}

Two generators fully meet two requirements each. \texttt{GutenTAG} derives a seed for each dataset from the base seed and the dataset's name, so its parallel output does not depend on the number of workers, but its output is indexed by position, with no frequency or calendar origin. TempoPFN stores a frequency, start and identity with every series and selects among 14 generators by weight, but its released command line seeds each run from the wall clock, and its per-generator seeds include Python's salted string hash, so a seed alone does not reproduce a run across interpreter launches. Four existing generators return a forecasting-ready panel: \texttt{gratis} through a \texttt{tsibble}, \texttt{GluonTS}, TempoPFN and \texttt{utilsforecast}. However, none of them combines this panel with perturbations that leave the base series untouched by design. A Python port of \texttt{gratis} in its repository, not yet released, returns bare arrays.

On other axes, other tools lead: \texttt{gratis} can target any user-specified vector of series features (its paper demonstrates up to 12) against our four, KernelSynth, ForecastPFN, TempoPFN and SarSim0 have each trained a published forecasting model, SarSim0 generates a billion series on the fly, and the multivariatizers of \citet{chronos2} impose multivariate structure that our joint generators do not attempt.

\section{Experimental protocol details}
\label{app:protocol}

\emph{Timestamps.} M4 series are distributed with an integer index, so we assign synthetic timestamps from a fixed origin at the frequency of the group. These carry no calendar meaning, and no model receives a calendar feature.

\emph{Seasonal periods.} The natural periods of Section~\ref{sec:protocol} serve both in the seasonal MASE denominator and in the cycle normalization. The M4 convention, which treats weekly and daily as non-seasonal, would leave too few frequencies with a defined period to distinguish the normalizations of Section~\ref{sec:crossover}.

\emph{Seeds.} Every random seed, for the synthetic pool and for initializing the pretrained, from-scratch and fine-tuned models, is derived from one fixed entropy value and a key naming its role, frequency, replicate and model, so no two replicates or roles share a seed. Two sharings are deliberate: both models pretrain on the same corpus within a (frequency, replicate), and every ablation arm reuses the control's draws (Section~\ref{sec:ablation}).

\emph{History levels.} Every history level satisfies two floors: it must admit at least one training window, and it must be at least twice the seasonal period, because seasonal MASE divides by the mean absolute difference at the seasonal lag and a history equal to the period supplies no such differences at all.

\emph{Replicates and run-to-run variation.} The grid has 30 replicates; the ablation arms run on the first 20. Training on a GPU need not be bit-reproducible, so 20 cells, four per frequency, were evaluated twice with identical seeds and checkpoints. Zero-shot forecasts were bit-identical, as no training happens at evaluation. The conditions trained at evaluation were bit-identical in all 16 non-hourly cells and differed in all four hourly ones; the largest difference in a cell mean was 0.12 replicate standard deviations for \texttt{from\_scratch}, below 0.01 for \texttt{pretrain\_ft}, and 0.52 for \texttt{from\_scratch\_long}. One AutoETS cell (daily) differed only by floating-point rounding, of the order of $10^{-13}$. Pretraining was not repeated, so its run-to-run variation is not measured. Run-to-run variation is part of the replicate-level variance that the tests already absorb, so it bounds the reproduction of point estimates rather than the inference.

\emph{Reproducibility.} Experiments install \texttt{SynForecast} 0.2.0 from PyPI, except the hourly follow-up of Section~\ref{sec:hourly}, which installs commit \texttt{b2338c8} from the repository. Every cell records the versions of each package it imported, its seeds and its realized step counts. The artifact at \artifacturl{} regenerates every table and figure.

\emph{Test construction.} Each history level is tested on its own 30 replicates; we do not pool across history levels, because the six levels within a replicate are nested truncations of the same subsample and pooling them would treat strongly dependent observations as independent. Correction is Holm within each comparison's family of 30 tests, and every significance statement above is post-correction.

\emph{Axis.} The crossover is located against the \emph{realized} mean training length of each cell, with full history included as an ordinary point. A nominal history level overstates what a cell trains on, because retaining the last $k$ observations returns all of a series shorter than $k$; at the largest weekly level the overstatement is 66 observations. The realized lengths were recomputed offline from the series subsamples.

\section{Pre-specification, outcomes and deviations}
\label{app:prespec}

The expectations and decision rules for the experiments of Section~\ref{sec:results} were written down and frozen on 2026-10-13, after smoke runs on reduced budgets had checked the pipeline and before the experiment was launched. The pre-specification is an internal document shipped with the artifact, not an external registration. It was written after earlier runs on the same evaluation subsamples, and several of its choices drew on them: the expectation that history relative to the horizon would be the least dispersed normalization and seasonal cycles the most, the exclusion of hourly from the ablation's primary population, and the log-ratio effect measure. The rule text, including the conclusion to be drawn under each crossover outcome, is in the artifact. Table~S7 in the supplementary material lists every expectation against its outcome.

The hourly follow-up of Section~\ref{sec:hourly} has its own pre-specification, written on 2026-10-30, after the main experiment and before the follow-up was launched. It was also written after a five-replicate pilot with NHITS alone, on a re-implementation of the pipeline, whose estimates it states. Its effect measure, tests and replication criterion are those of the main experiment. The last four rows of Table~S7 score it.

\emph{Deviations.} Four departures from the pre-specification bear on what is reported.
\begin{enumerate}
  \item \emph{Ablation replicates.} The ablation was fixed at replicates 0--9 and run on 0--19. The change was decided before launch, because ten replicate clusters give under-covering bootstrap intervals; the runner whose hash was recorded at launch already uses 20. Table~\ref{tab:ablation} reports both; on the pre-specified replicates only the classical family replicates across models.
  \item \emph{Crossover axis.} The rule as first fixed located the crossover on the nominal history levels with full history excluded. After an earlier run, in which that placement excluded the crossover itself at monthly and quarterly, we moved to realized training lengths with full history included (\ref{app:protocol}). On the original axis, this run's verdict is unchanged -- no collapse for either model -- but only daily and weekly cross, so the coefficients of variation rest on two frequencies: history relative to the horizon is least dispersed (NHITS 0.17, LightGBM 0.19), only 1.30 and 1.26 times clear of the runner-up against the required 2.
  \item \emph{Budget check.} With ten replicates, the pre-specified Holm family of 30 tests cannot reach significance, so the check is reported as estimates with intervals (Section~\ref{sec:crossover}; Table~S5 in the supplementary material).
  \item \emph{Run-to-run noise.} The from-scratch shifts of 0.12 replicate standard deviations, and 0.52 with the 1{,}000-step budget, exceed the pre-specified 0.1, so differences of that size between trained conditions are within run-to-run noise (\ref{app:protocol}).
\end{enumerate}
The analyses that report the ablation on both replicate sets, the budget check as intervals, the verdict on the original axis, the zero-shot/Seasonal Naive ratios and run-to-run noise by frequency were added in revision and are not pre-specified; they are in the artifact as a separate script.

\section{Generation cost}
\label{app:cost}

\emph{Throughput against worker count.} We hold the total work fixed at 864 series of 1{,}024 observations from the 54-instance pretraining pool and vary the worker count. On a machine with 20 logical cores (Linux under WSL2), throughput rises from 171 series per second with one worker to a plateau of 370 to 420 between six and twelve workers, peaks at eight with a speedup of 2.45, and falls to 321 at 20 (Table~\ref{tab:throughput}a). A random walk alone generates 34{,}000 to 40{,}000 series per second at any worker count, so for cheap mechanisms the cost of a series is fixed per-call work. The pool is limited instead by its three KernelSynth instances, which take about four-fifths of single-worker time and factorize a dense covariance matrix with a multithreaded linear-algebra library, so that worker and library threads compete for cores beyond the plateau. The library's default of using every core is therefore not the best setting, and a caller generating at scale should set the worker count explicitly.

\emph{Cost across mechanisms.} On one series of length 1{,}000, the public generation call costs a median of 0.87~ms across the 32 generators, mostly fixed per-call overhead that amortizes in batch; KernelSynth costs 101 times the median, and without it the spread is sixfold.

\emph{Cost concentrates in length.} At 9{,}919 observations, the longest daily series, one KernelSynth series takes 8.2 seconds and 5.7~GB of resident memory (97\% of the summed cost of one series from each of the 19 classes in the pool), and its cost grows superlinearly with length. Table~\ref{tab:throughput}b shows the consequence for our experiment: the quarterly draw (lengths 16--866) takes 1.3 seconds and 0.3~GB, the daily draw (93--9{,}919) 391 seconds and 22.7~GB. Generating each (frequency, replicate) draw once and sharing it across both models and all ablation arms kept the experiment's 150 draws to 2.5 CPU-container-hours, against 8.4 GPU-hours for the 1{,}730 pretraining runs.

\begin{widetable}[tbp]
\centering
\caption{Generation throughput. Panel (a) holds total work fixed at 864 series of 1{,}024 observations from the 54-instance pretraining pool and varies the worker count; medians of five interleaved repetitions on 20 logical cores. Panel (b) is the cost of one pretraining draw at each frequency, as the experiments of this section generated it: 54 instances, 26 series each, four workers. Each draw runs in a fresh process, and peak resident memory is that process's own.}
\label{tab:throughput}
\begin{tabular}{lrrrr}
\multicolumn{5}{l}{\emph{(a) Throughput against worker count}} \\
\toprule
Workers & Median (s) & Series/s & Speedup & Efficiency \\
\midrule
1  & 5.05 & 171 & 1.00 & 1.00 \\
2  & 3.89 & 222 & 1.30 & 0.65 \\
4  & 2.77 & 312 & 1.82 & 0.46 \\
6  & 2.33 & 371 & 2.17 & 0.36 \\
8  & 2.06 & \textbf{419} & \textbf{2.45} & 0.31 \\
12 & 2.27 & 381 & 2.23 & 0.19 \\
20 & 2.69 & 321 & 1.87 & 0.09 \\
\bottomrule
\end{tabular}

\vspace{1.5ex}

\begin{tabular}{lrrrrr}
\multicolumn{6}{l}{\emph{(b) Cost of one pretraining draw, by frequency}} \\
\toprule
Frequency & Lengths & Rows & Seconds & Series/s & Peak RSS \\
\midrule
Quarterly & 16--866 & 627{,}364 & 1.3 & 1069 & 0.31~GB \\
Hourly    & 700--960 & 1{,}167{,}877 & 3.6 & 388 & 0.46~GB \\
Weekly    & 80--2{,}597 & 1{,}908{,}521 & 12.1 & 116 & 1.95~GB \\
Monthly   & 42--2{,}794 & 2{,}012{,}590 & 14.3 & 98 & 2.13~GB \\
Daily     & 93--9{,}919 & 7{,}120{,}903 & 391.3 & 4 & 22.67~GB \\
\bottomrule
\end{tabular}
\end{widetable}

\section{Overlay orthogonality}
\label{app:overlaytest}

Both invariance properties are asserted by the test suite. One test checks that a given seed produces identical values with one and with four workers, for three batch-path generators and a composed pool. Another test checks orthogonality for all 32 generators, using the flag columns that report which observations each overlay touched: every untouched observation is bit-identical to the run without the overlay, series lengths are unchanged, and each overlay lands in the same place whatever other overlays are enabled. For changepoints, which alter everything after them, the test checks the observations before the first changepoint, and a level-shift anomaly, flagged at its start only, is checked over its configured duration.

The orthogonality test exists because the property did not hold everywhere. Before version 0.2.0, the generators outside the batch path (KernelSynth, copula, VAR, multivariate TCM, IoT sensor and daily active users) drew all three overlays from one stream, so enabling one overlay could move another. Version 0.2.0 gives every series three perturbation seeds on every path and adds the test to guard against this failure.

\section{Ablation results}
\label{app:results}

Table~\ref{tab:ablation} reports the ablation of Section~\ref{sec:ablation} on both replicate sets, the run as executed and the pre-specified one, together with the hourly effects that Figure~\ref{fig:ablation} omits.

\begin{widetable}[tbp]
\centering
\caption{Mechanism-family ablation. Effect $= 100 \times \log$(geometric-mean zero-shot MASE, arm / control); positive means removing the family made zero-shot accuracy \emph{worse}, so the family contributes. For each family, the first row is the run as executed (replicates 0--19) and the second the pre-specified replicates 0--9 (\ref{app:prespec}). Primary population: daily, weekly, monthly and quarterly, three history levels each; 95\% interval from a bootstrap over replicates; bold where the interval excludes zero. Hourly is reported separately, as a point estimate on replicates 0--19. Families: classical = SARIMA, ETS, INAR; Gaussian process = four GP instances; heavy-tail and regime = GARCH, L\'evy, regime switching; other stochastic = fractional Brownian motion, chaotic systems, bounded processes, cyclic; domain = intermittent demand, energy load, IoT sensor, vital signs, of whose 11 instances one almost never contributes (Section~\ref{sec:protocol}); procedural = TSI, TCM, KernelSynth; MAR = three non-seasonal mixture-autoregressive instances.}
\label{tab:ablation}
\footnotesize\setlength{\tabcolsep}{3pt}
\begin{tabular}{llrllrr}
\toprule
& & & \multicolumn{2}{c}{Primary population} & \multicolumn{2}{c}{Hourly} \\
\cmidrule(lr){4-5}\cmidrule(lr){6-7}
Family removed & Replicates & Inst. & NHITS & LightGBM & NHITS & LightGBM \\
\midrule
Classical & 0--19 & 11 & \textbf{\boldmath$+1.83$ $[+1.15, +2.51]$} & \textbf{\boldmath$+2.30$ $[+1.93, +2.63]$} & $+2.2$ & $-2.3$ \\
 & 0--9 & & \textbf{\boldmath$+1.69$ $[+0.91, +2.51]$} & \textbf{\boldmath$+2.46$ $[+1.85, +2.98]$} & & \\
Gaussian process & 0--19 & 4 & \textbf{\boldmath$+0.88$ $[+0.25, +1.53]$} & \textbf{\boldmath$+0.93$ $[+0.71, +1.13]$} & $+9.8$ & $+10.8$ \\
 & 0--9 & & $+0.90$ $[-0.18, +2.06]$ & \textbf{\boldmath$+0.98$ $[+0.67, +1.27]$} & & \\
Heavy-tail, regime & 0--19 & 6 & \textbf{\boldmath$+0.98$ $[+0.47, +1.45]$} & \textbf{\boldmath$+0.83$ $[+0.52, +1.11]$} & $-1.0$ & $-6.6$ \\
 & 0--9 & & $+0.81$ $[-0.04, +1.60]$ & \textbf{\boldmath$+0.86$ $[+0.49, +1.26]$} & & \\
Other stochastic & 0--19 & 10 & \textbf{\boldmath$-0.83$ $[-1.47, -0.22]$} & \textbf{\boldmath$-0.79$ $[-1.04, -0.52]$} & $-13.5$ & $-1.0$ \\
 & 0--9 & & $-0.66$ $[-1.61, +0.22]$ & \textbf{\boldmath$-0.80$ $[-1.18, -0.41]$} & & \\
Domain & 0--19 & 11 & $+0.49$ $[-0.11, +1.09]$ & $-0.25$ $[-0.54, +0.06]$ & $+7.6$ & $+19.2$ \\
 & 0--9 & & $+0.26$ $[-0.61, +1.17]$ & $-0.20$ $[-0.64, +0.22]$ & & \\
Procedural & 0--19 & 9 & $-0.35$ $[-1.15, +0.43]$ & \textbf{\boldmath$-0.48$ $[-0.77, -0.21]$} & $+4.9$ & $+6.1$ \\
 & 0--9 & & $-0.48$ $[-1.87, +0.86]$ & \textbf{\boldmath$-0.48$ $[-0.93, -0.07]$} & & \\
MAR & 0--19 & 3 & $-0.05$ $[-0.66, +0.54]$ & \textbf{\boldmath$-0.61$ $[-0.80, -0.42]$} & $+0.3$ & $0.0$ \\
 & 0--9 & & $+0.05$ $[-0.72, +0.73]$ & \textbf{\boldmath$-0.54$ $[-0.79, -0.27]$} & & \\
\bottomrule
\end{tabular}
\end{widetable}

\end{document}


\begin{frontmatter}
\title{Supplementary material for ``SynForecast: Controlled Synthetic Time Series for Forecasting Research''}
\author[aff1]{Olivier Sprangers\corref{cor1}}
\ead{olivier@nixtla.io}
\author[aff1]{Cristian Challu}
\author[aff1]{Jan Rathfelder}
\author[aff1]{Han Wang}
\author[aff1]{Max Mergenthaler Canseco}
\cortext[cor1]{Corresponding author.}
\affiliation[aff1]{organization={Nixtla},
  city={San Francisco},
  state={CA},
  country={United States}}
\end{frontmatter}

Table~\ref{tab:sources} gives, for each of the 32 generators in synforecast 0.2.0, the model each generator implements and the work that introduced it. The five domain simulators have no canonical source: they compose stylized components (daily and weekly profiles, drifts, events, failures) and their tests pin those designed behaviors rather than a published model's statistics. The random walk, the seasonal sinusoid, the cyclic generator and the Poisson process are textbook constructions and are listed without a source.

{\footnotesize\setlength{\tabcolsep}{4pt}
\begin{xltabular}{\textwidth}{@{}l>{\raggedright\arraybackslash}X>{\raggedright\arraybackslash}p{0.34\textwidth}@{}}
\caption{Generators in synforecast 0.2.0 (class names without the \texttt{Generator} suffix), grouped as in Section~3 of the article, with the model each implements and its source. A dash marks a textbook construction or a stylized domain simulator.}\label{tab:sources} \\
\toprule
Generator & Model & Source \\
\midrule
\endfirsthead
\multicolumn{3}{@{}l}{\emph{Table~\ref{tab:sources}, continued}} \\
\toprule
Generator & Model & Source \\
\midrule
\endhead
\multicolumn{3}{@{}l}{\emph{Statistical}} \\
\texttt{RandomWalk} & Random walk with drift & -- \\
\texttt{Seasonal} & Sinusoid plus linear trend and noise & -- \\
\texttt{SARIMA} & Seasonal ARIMA $(p,d,q)\times(P,D,Q)_s$ with optional regressors & \citealp{box2008time} \\
\texttt{ETS} & Innovations state space form of exponential smoothing & \citealp{hyndman2002state,hyndman2008forecasting} \\
\texttt{INAR} & INAR($p$) with binomial thinning, Poisson or negative binomial innovations & \citealp{steutel1979discrete,alosh1987first,du1991integer} \\
\addlinespace
\multicolumn{3}{@{}l}{\emph{Stochastic}} \\
\texttt{GARCH} & GARCH($p$, $q$) & \citealp{bollerslev1986garch} \\
\texttt{OrnsteinUhlenbeck} & Mean-reverting diffusion, Euler--Maruyama & \citealp{uhlenbeck1930theory} \\
\texttt{GeometricBrownianMotion} & Log-normal diffusion, exact discretization & \citealp{black1973pricing} \\
\texttt{JumpDiffusion} & Geometric Brownian motion with compound Poisson log-normal jumps & \citealp{merton1976option} \\
\texttt{PoissonProcess} & Homogeneous Poisson counts & -- \\
\texttt{Cyclic} & Superposed sinusoids with slowly modulated frequency & -- \\
\texttt{FractionalBrownianMotion} & Fractional Brownian motion; circulant embedding by default & \citealp{mandelbrot1968fractional,davies1987tests} \\
\texttt{HawkesProcess} & Self-exciting point process, exponential or power-law kernel & \citealp{hawkes1971spectra} \\
\texttt{StochasticVolatility} & Heston; SABR & \citealp{heston1993closed,hagan2002managing} \\
\texttt{RegimeSwitching} & Markov-switching AR(1) & \citealp{hamilton1989new} \\
\texttt{ChaoticSystem} & Lorenz system; logistic map; Mackey--Glass equation & \citealp{lorenz1963deterministic,may1976simple,mackey1977oscillation} \\
\texttt{BoundedProcess} & Beta autoregression; logit-normal AR(1) & \citealp{rocha2009beta} \\
\texttt{LevyProcess} & $\alpha$-stable increments, Chambers--Mallows--Stuck sampling & \citealp{chambers1976method} \\
\addlinespace
\multicolumn{3}{@{}l}{\emph{Multivariate}} \\
\texttt{Copula} & Gaussian or $t$ copula with arbitrary marginals & \citealp{nelsen2006introduction,demarta2005t} \\
\texttt{VAR} & Stable vector autoregression VAR($p$) & \citealp{lutkepohl2005new} \\
\texttt{GaussianProcess} & Gaussian-process prior; RBF, Mat\'ern or periodic kernel & \citealp{rasmussen2006gaussian} \\
\addlinespace
\multicolumn{3}{@{}l}{\emph{Domain}} \\
\texttt{IntermittentDemand} & Bernoulli demand occurrence times a demand-size distribution & \citealp{croston1972forecasting} \\
\texttt{StateSpace} & Linear Gaussian state space model, or user-supplied dynamics & \citealp{durbin2012time} \\
\texttt{EnergyLoad} & Stylized: daily, weekly and yearly profiles, temperature response & -- \\
\texttt{IoTSensor} & Stylized: drift, battery degradation, sensor failures & -- \\
\texttt{VitalSigns} & Stylized: baseline, circadian rhythm, physiological events & -- \\
\texttt{Clickstream} & Stylized: Poisson sessions, diurnal profile, bot traffic & -- \\
\texttt{DailyActiveUsers} & Stylized: growth, weekend effect, decaying event boosts & -- \\
\addlinespace
\multicolumn{3}{@{}l}{\emph{Pretraining}} \\
\texttt{TSI} & Random trend, seasonality and irregular composition & \citealp{bahrpeyma2021methodology} \\
\texttt{TCM} & Random temporal structural causal model & \citealp{runge2023causal} \\
\texttt{KernelSynth} & Gaussian-process prior over randomly composed kernels & \citealp{ansari2024chronos} \\
\texttt{MAR} & Mixture autoregression with feature-targeted search & \citealp{wongli2000mar,kang2020gratis} \\
\bottomrule
\end{xltabular}}

For TSI and TCM, the pair that \citet{chronos2} use as univariate generators, the cited works supply the construction rather than the generator: TSI composes components after \citet{bahrpeyma2021methodology} with its own component families and sampling distributions, and TCM follows the temporal structural-causal framing of \citet{runge2023causal} with its own graph sampler and edge functions.

\section*{Detailed results}

Table~\ref{tab:fig3ci} gives every value of Figure~3 of the article with its uncertainty, and Table~\ref{tab:overlay} the values plotted in Figure~6 of the article, with the error of each method on the unperturbed panels for scale. Tables~\ref{tab:tost}, \ref{tab:e1b} and~\ref{tab:zsratio} give the fine-tuning tests of Section~4.3 of the article, its budget check of Section~4.4, and the zero-shot error relative to Seasonal Naive on which the reading of the crossover in Section~4.4 rests. Table~\ref{tab:scorecard} scores every pre-specified expectation against its outcome, as Appendix~C of the article describes.

\begin{table}[tbp]
\centering
\caption{The values of Figure~3 of the article with their uncertainty: relative improvement (\%) of the condition named first over the condition named second, NHITS, mean over 30 replicates, with the half-width of the 95\% $t$ interval over replicates; history levels from the shortest (left) to full history (right), as in Figure~3 of the article.}
\label{tab:fig3ci}
\footnotesize\setlength{\tabcolsep}{3pt}
\begin{tabular}{@{}lrrrrrr@{}}
\toprule
Frequency & \multicolumn{6}{c}{History level} \\
\midrule
\multicolumn{7}{@{}l}{\emph{(a) \texttt{zero\_shot} over \texttt{from\_scratch}}} \\
Quarterly & $+$32.5\,{\tiny$\pm$0.9} & $+$33.4\,{\tiny$\pm$0.7} & $+$31.5\,{\tiny$\pm$1.3} & $+$29.1\,{\tiny$\pm$1.6} & $+$21.0\,{\tiny$\pm$2.0} & $-$7.3\,{\tiny$\pm$2.1} \\
Monthly & $+$26.1\,{\tiny$\pm$1.0} & $+$24.5\,{\tiny$\pm$1.1} & $+$24.3\,{\tiny$\pm$1.1} & $+$19.8\,{\tiny$\pm$1.2} & $+$10.5\,{\tiny$\pm$1.8} & $-$8.9\,{\tiny$\pm$1.8} \\
Weekly & $+$6.5\,{\tiny$\pm$0.9} & $+$3.6\,{\tiny$\pm$0.8} & $+$2.2\,{\tiny$\pm$1.0} & $-$0.5\,{\tiny$\pm$1.1} & $-$7.9\,{\tiny$\pm$1.3} & $-$17.2\,{\tiny$\pm$3.6} \\
Daily & $+$34.1\,{\tiny$\pm$0.9} & $+$29.4\,{\tiny$\pm$1.0} & $+$17.5\,{\tiny$\pm$1.4} & $+$8.8\,{\tiny$\pm$1.3} & $-$1.0\,{\tiny$\pm$1.7} & $-$17.4\,{\tiny$\pm$2.8} \\
Hourly & $-$51.0\,{\tiny$\pm$5.8} & $-$60.5\,{\tiny$\pm$5.7} & $-$82.2\,{\tiny$\pm$7.0} & $-$122.8\,{\tiny$\pm$8.7} & $-$248.0\,{\tiny$\pm$13.4} & $-$358.6\,{\tiny$\pm$19.7} \\
\addlinespace
\multicolumn{7}{@{}l}{\emph{(b) \texttt{pretrain\_ft} over \texttt{from\_scratch}}} \\
Quarterly & $+$15.3\,{\tiny$\pm$1.5} & $+$14.7\,{\tiny$\pm$1.6} & $+$15.2\,{\tiny$\pm$1.4} & $+$15.3\,{\tiny$\pm$1.9} & $+$13.8\,{\tiny$\pm$2.1} & $+$3.3\,{\tiny$\pm$1.5} \\
Monthly & $+$6.5\,{\tiny$\pm$1.0} & $+$9.1\,{\tiny$\pm$0.8} & $+$6.3\,{\tiny$\pm$0.9} & $+$4.5\,{\tiny$\pm$1.0} & $+$2.9\,{\tiny$\pm$1.2} & $+$0.6\,{\tiny$\pm$1.1} \\
Weekly & $+$3.7\,{\tiny$\pm$0.8} & $+$3.0\,{\tiny$\pm$0.6} & $+$1.9\,{\tiny$\pm$0.8} & $+$1.1\,{\tiny$\pm$0.9} & $-$1.9\,{\tiny$\pm$0.6} & $-$0.8\,{\tiny$\pm$1.0} \\
Daily & $+$4.3\,{\tiny$\pm$0.5} & $+$5.9\,{\tiny$\pm$0.9} & $+$4.2\,{\tiny$\pm$1.3} & $+$3.8\,{\tiny$\pm$1.0} & $+$3.1\,{\tiny$\pm$1.2} & $-$1.3\,{\tiny$\pm$0.9} \\
Hourly & $-$10.3\,{\tiny$\pm$2.1} & $-$11.1\,{\tiny$\pm$1.7} & $-$13.5\,{\tiny$\pm$2.3} & $-$15.3\,{\tiny$\pm$2.3} & $-$25.2\,{\tiny$\pm$2.7} & $-$26.4\,{\tiny$\pm$3.0} \\
\addlinespace
\multicolumn{7}{@{}l}{\emph{(c) \texttt{pretrain\_ft} over \texttt{zero\_shot}}} \\
Quarterly & $-$25.5\,{\tiny$\pm$2.5} & $-$28.2\,{\tiny$\pm$2.7} & $-$24.0\,{\tiny$\pm$2.2} & $-$19.6\,{\tiny$\pm$1.8} & $-$9.2\,{\tiny$\pm$1.8} & $+$9.8\,{\tiny$\pm$1.4} \\
Monthly & $-$26.6\,{\tiny$\pm$1.3} & $-$20.6\,{\tiny$\pm$1.3} & $-$23.9\,{\tiny$\pm$1.4} & $-$19.3\,{\tiny$\pm$1.4} & $-$8.6\,{\tiny$\pm$1.3} & $+$8.7\,{\tiny$\pm$1.1} \\
Weekly & $-$3.0\,{\tiny$\pm$0.6} & $-$0.7\,{\tiny$\pm$0.6} & $-$0.3\,{\tiny$\pm$0.6} & $+$1.5\,{\tiny$\pm$0.7} & $+$5.5\,{\tiny$\pm$1.1} & $+$13.5\,{\tiny$\pm$2.5} \\
Daily & $-$45.3\,{\tiny$\pm$2.0} & $-$33.4\,{\tiny$\pm$1.4} & $-$16.2\,{\tiny$\pm$1.2} & $-$5.7\,{\tiny$\pm$1.1} & $+$4.0\,{\tiny$\pm$1.1} & $+$13.4\,{\tiny$\pm$1.8} \\
Hourly & $+$26.4\,{\tiny$\pm$2.3} & $+$30.3\,{\tiny$\pm$2.1} & $+$37.2\,{\tiny$\pm$2.0} & $+$47.8\,{\tiny$\pm$1.7} & $+$63.7\,{\tiny$\pm$1.3} & $+$72.2\,{\tiny$\pm$0.9} \\
\bottomrule
\end{tabular}
\end{table}

\begin{table}[tbp]
\centering
\caption{Change in MASE caused by one level shift, by its position relative to the forecast origin; mean over 5 replicates of 300 monthly SARIMA series (standard error across replicates). Breaks ``before'' fall in the training window; ``after'' falls in the forecast window. MASE without a break: Seasonal Naive 1.012, AutoETS 0.807, NHITS 0.716.}
\label{tab:overlay}
\begin{tabular}{lrrr}
\toprule
Break position (obs) & Seas.\ Naive & AutoETS & NHITS \\
\midrule
72 before & $+0.000$\,(0.000) & $+0.070$\,(0.006) & $+0.005$\,(0.002) \\
36 before & $+0.000$\,(0.000) & $+0.070$\,(0.005) & $+0.014$\,(0.004) \\
18 before & $+0.000$\,(0.000) & $+0.078$\,(0.004) & $+0.334$\,(0.229) \\
6 before & $+1.140$\,(0.105) & $+0.150$\,(0.024) & $+1.393$\,(0.037) \\
1 before & $+1.620$\,(0.162) & $+1.441$\,(0.127) & $+1.903$\,(0.107) \\
6 after & $+1.140$\,(0.110) & $+1.280$\,(0.086) & $+1.333$\,(0.114) \\
\bottomrule
\end{tabular}
\end{table}

\begin{table}[tbp]
\centering
\caption{\texttt{pretrain\_ft} against \texttt{from\_scratch} and against \texttt{zero\_shot}, NHITS, 30 replicates. Positive favors fine-tuning. Each history level is tested on its 30 replicates with a Wilcoxon signed-rank test; ``Levels'' counts the six levels that reach significance after Holm correction within each family of 30 tests (5 frequencies $\times$ 6 history levels), one family per comparison. Mean and win rate are averaged over levels; $\dagger$ marks a frequency whose per-level effect changes sign with history, for which no mean is given (per-level values with intervals are in Table~S2). MDE is the minimum detectable effect at 80\% power on the per-replicate percentage differences, under a normal approximation, as a range across levels.}
\label{tab:tost}
\small\setlength{\tabcolsep}{3pt}
\begin{tabular}{lrrrrrrr}
\toprule
& \multicolumn{4}{c}{vs \texttt{from\_scratch}} & \multicolumn{3}{c}{vs \texttt{zero\_shot}} \\
\cmidrule(lr){2-5}\cmidrule(lr){6-8}
Frequency & Mean (\%) & Win rate & Levels & MDE (\%) & Mean (\%) & Win rate & Levels \\
\midrule
Quarterly & $+12.96$ & 0.97 & 6/6 & 2.0--2.9 & $\dagger$ & 0.17 & 6/6 \\
Monthly   & $+4.98$ & 0.91 & 5/6 & 1.1--1.6 & $\dagger$ & 0.17 & 6/6 \\
Weekly    & $\dagger$ & 0.68 & 5/6 & 0.8--1.4 & $\dagger$ & 0.60 & 4/6 \\
Daily     & $\dagger$ & 0.82 & 6/6 & 0.7--1.7 & $\dagger$ & 0.32 & 6/6 \\
Hourly    & $-16.96$ & 0.01 & 6/6 & 2.3--4.1 & $+46.29$ & 1.00 & 6/6 \\
\bottomrule
\end{tabular}
\end{table}

\begin{table}[tbp]
\centering
\caption{Sensitivity check E1b: improvement (\%) of NHITS trained from scratch for 1{,}000 steps over 300 steps, mean over the ten replicates that ran it, by frequency and history level (nominal history level in parentheses). Positive favors 1{,}000 steps. Bold where the 95\% replicate-bootstrap interval excludes zero. With ten replicates and Holm correction over 30 tests no level can reach significance (smallest attainable adjusted $p = 0.059$), so the check is reported as estimates.}
\label{tab:e1b}
\footnotesize\setlength{\tabcolsep}{3pt}
\begin{tabular}{lrrrrrr}
\toprule
Frequency & \multicolumn{6}{c}{History level} \\
\midrule
Quarterly & \textbf{$-$2.4}\,\tiny{(16)} & \textbf{$-$3.3}\,\tiny{(20)} & \textbf{$-$3.6}\,\tiny{(32)} & \textbf{$-$5.3}\,\tiny{(48)} & $-$3.1\,\tiny{(64)} & $+$1.8\,\tiny{(full)} \\
Monthly & \textbf{$-$2.1}\,\tiny{(36)} & \textbf{$-$3.4}\,\tiny{(48)} & \textbf{$-$1.9}\,\tiny{(72)} & $-$0.4\,\tiny{(108)} & $-$1.3\,\tiny{(144)} & $+$2.2\,\tiny{(full)} \\
Weekly & \textbf{$-$2.3}\,\tiny{(104)} & \textbf{$-$1.7}\,\tiny{(156)} & \textbf{$-$2.4}\,\tiny{(208)} & $-$1.2\,\tiny{(260)} & $+$0.5\,\tiny{(416)} & $+$0.9\,\tiny{(full)} \\
Daily & \textbf{$+$1.9}\,\tiny{(28)} & $-$1.6\,\tiny{(56)} & \textbf{$-$3.0}\,\tiny{(112)} & \textbf{$-$3.9}\,\tiny{(182)} & $+$0.3\,\tiny{(364)} & \textbf{$-$1.3}\,\tiny{(full)} \\
Hourly & $+$3.0\,\tiny{(96)} & \textbf{$+$17.7}\,\tiny{(144)} & \textbf{$+$19.7}\,\tiny{(240)} & \textbf{$+$22.6}\,\tiny{(384)} & \textbf{$+$21.9}\,\tiny{(672)} & \textbf{$+$15.6}\,\tiny{(full)} \\
\bottomrule
\end{tabular}
\end{table}

\begin{table}[tbp]
\centering
\caption{Mean zero-shot MASE relative to mean Seasonal Naive MASE over 30 replicates, at the shortest and at full history, control arm.}
\label{tab:zsratio}
\footnotesize
\begin{tabular}{lrrrr}
\toprule
& \multicolumn{2}{c}{NHITS} & \multicolumn{2}{c}{LightGBM} \\
\cmidrule(lr){2-3}\cmidrule(lr){4-5}
Frequency & Shortest & Full & Shortest & Full \\
\midrule
Quarterly & 1.12 & 1.08 & 1.11 & 1.17 \\
Monthly & 1.07 & 1.02 & 1.01 & 0.99 \\
Weekly & 0.67 & 0.56 & 0.69 & 0.68 \\
Daily & 0.95 & 0.94 & 0.93 & 1.12 \\
Hourly & 2.41 & 5.40 & 3.13 & 6.37 \\
\bottomrule
\end{tabular}
\end{table}

\begin{table}[tbp]
\centering
\caption{Pre-specified expectations against outcomes.}
\label{tab:scorecard}
\footnotesize
\begin{tabularx}{\textwidth}{@{}l>{\raggedright\arraybackslash}X>{\raggedright\arraybackslash}X l@{}}
\toprule
Item & Expectation & Outcome & Held \\
\midrule
Baselines & A statistical baseline is lowest in at least 8 of the 10 rows of Table~1 of the article & 9 of 10 & Yes \\
Crossover & Zero-shot beats scratch at the shortest history and loses at full history at four of five frequencies, hourly negative; the rule returns no collapse, history/horizon least and cycles most dispersed & As expected, for both models & Yes \\
Fine-tuning & Fine-tuning beats scratch at quarterly at all six levels, at least four significant & Six of six significant, $+13$\% & Yes \\
 & The quarterly effect is presented as reliable only if fine-tuning is nowhere significantly worse than zero-shot & Significantly worse at short histories & Not presented \\
 & The sign alternation of an earlier run at daily and hourly does not recur & It does not & Yes \\
Budget (E1b) & 1{,}000-step scratch beats 300 at full history at four or more frequencies; the zero-shot advantage survives & Better at full history at hourly; at the other frequencies within $\pm 2.2\%$, and worse at most shorter histories; the advantage survives & No \\
Ablation & No family replicates across both models outside hourly, on replicates 0--9 & Classical replicates on 0--9; four families on 0--19 & No \\
Hourly scaler & The window scaler stops the zero-shot/Seasonal Naive ratio growing (at most 1.2 times) & 0.88 times & Yes \\
Coverage & Hourly coverage gap above 0.50; below 0.30 elsewhere & 0.48; 0.07 to 0.26 elsewhere & Partly \\
Run-to-run & Zero-shot bit-identical; trained-condition shifts below 0.1 replicate standard deviations & Zero-shot identical; shifts up to 0.12 (scratch) and 0.52 (scratch, 1{,}000 steps), all at hourly & Partly \\
\midrule
Cycles (follow-up) & Calendar cycles lower hourly zero-shot error at all six levels for NHITS, and the effect replicates on LightGBM & By 43 to 62\% (NHITS) and 52 to 63\% (LightGBM); replicates at six of six levels & Yes \\
 & NHITS with cycles beats scratch at 96 observations and loses at full history, crossing between 240 and 672 & $+16$\%, $-70$\%; crossing at 323 & Yes \\
 & Zero-shot with cycles stays worse than Seasonal Naive at every level, its ratio growing less than 1.5-fold & 1.34 to 2.01 times; 1.503-fold & Partly \\
 & The effect of the cycles is smaller under the window scaler at every level & 26 to 29\% against 43 to 62\% & Yes \\
\bottomrule
\end{tabularx}
\end{table}

\bibliographystyle{elsarticle-harv}
\bibliography{supplementary,references}